\documentclass[sigconf,authordraft,review=false]{acmart}
\renewcommand\footnotetextcopyrightpermission[1]{} 
\usepackage{fancyhdr}
\usepackage{threeparttable}
\usepackage{todonotes}
\usepackage{multirow}
\usepackage{tabularx}
\usepackage{graphicx} 
\usepackage{caption}
\usepackage{subcaption} 
\usepackage{adjustbox}
\usepackage{booktabs}
\usepackage{colortbl}  %
\usepackage{xcolor}
\usepackage{hyperref}
\usepackage{algorithm}
\usepackage{algorithmic}
\usepackage{lineno} 

\AtBeginDocument{%
  \providecommand\BibTeX{{%
    \normalfont B\kern-0.5em{\scshape i\kern-0.25em b}\kern-0.8em\TeX}}}

\begin{document}

\title{Discover Your Neighbors: Advanced Stable Test-Time Adaptation in Dynamic World}


\author{Qinting Jiang,}
\affiliation{%
  \institution{Tsinghua University}  
  \city{Shenzhen}
  \country{China}
}
\email{jqt23@mails.tsinghua.edu.cn}

\author{Chuyang Ye}
\affiliation{%
  \institution{Shenzhen Technology University}  
  \city{Shenzhen}
  \country{China}
}
\email{youngyorkye@gmail.com}

\author{Dongyan Wei}
\affiliation{%
  \institution{Shenzhen Technology University}  
  \city{Shenzhen}
  \country{China}
}
\email{Dongyanwei7@outlook.com}
\author{Yuan Xue}
\affiliation{%
  \institution{Tsinghua University}  
  \city{Shenzhen}
  \country{China}
}
\email{xuey21@mails.tsinghua.edu.cn}

\author{Jingyan Jiang}
\affiliation{%
  \institution{Shenzhen Technology University}  
  \city{Shenzhen}
  \country{China}
}
\email{jiangjingyan@sztu.edu.cn}
\author{Zhi Wang
}
\affiliation{%
  \institution{Tsinghua University}  
  \city{Shenzhen}
  \country{China}
}
\email{wangzhi@sz.tsinghua.edu.cn}





\begin{abstract}

Despite progress, deep neural networks still suffer performance declines under distribution shifts between training and test domains,  leading to a substantial decrease in Quality of Experience (QoE) for multimedia applications. Existing test-time adaptation (TTA) methods are challenged by dynamic, multiple test distributions within batches. This work provides a new perspective on analyzing batch normalization techniques through class-related and class-irrelevant features, our observations reveal combining source and test batch normalization statistics robustly characterizes target distributions. However, test statistics must have high similarity. We thus propose \textbf{D}iscover \textbf{Y}our \textbf{N}eighbours (DYN), the first backward-free approach specialized for dynamic TTA. The core innovation is identifying similar samples via instance normalization statistics and clustering into groups which provides consistent class-irrelevant representations. Specifically, Our DYN consists of layer-wise instance statistics clustering (LISC) and cluster-aware batch normalization (CABN). In LISC, we perform layer-wise clustering of approximate feature samples at each BN layer by calculating the cosine similarity of instance normalization statistics across the batch. CABN then aggregates SBN and TCN statistics to collaboratively characterize the target distribution, enabling more robust representations. Experimental results validate DYN's robustness and effectiveness, demonstrating maintained performance under dynamic data stream patterns.
\end{abstract}    






\maketitle

\section{Introduction}
\label{sec:intro}

\begin{figure*}[htbp]   
\centering
\includegraphics[width=0.8\textwidth]{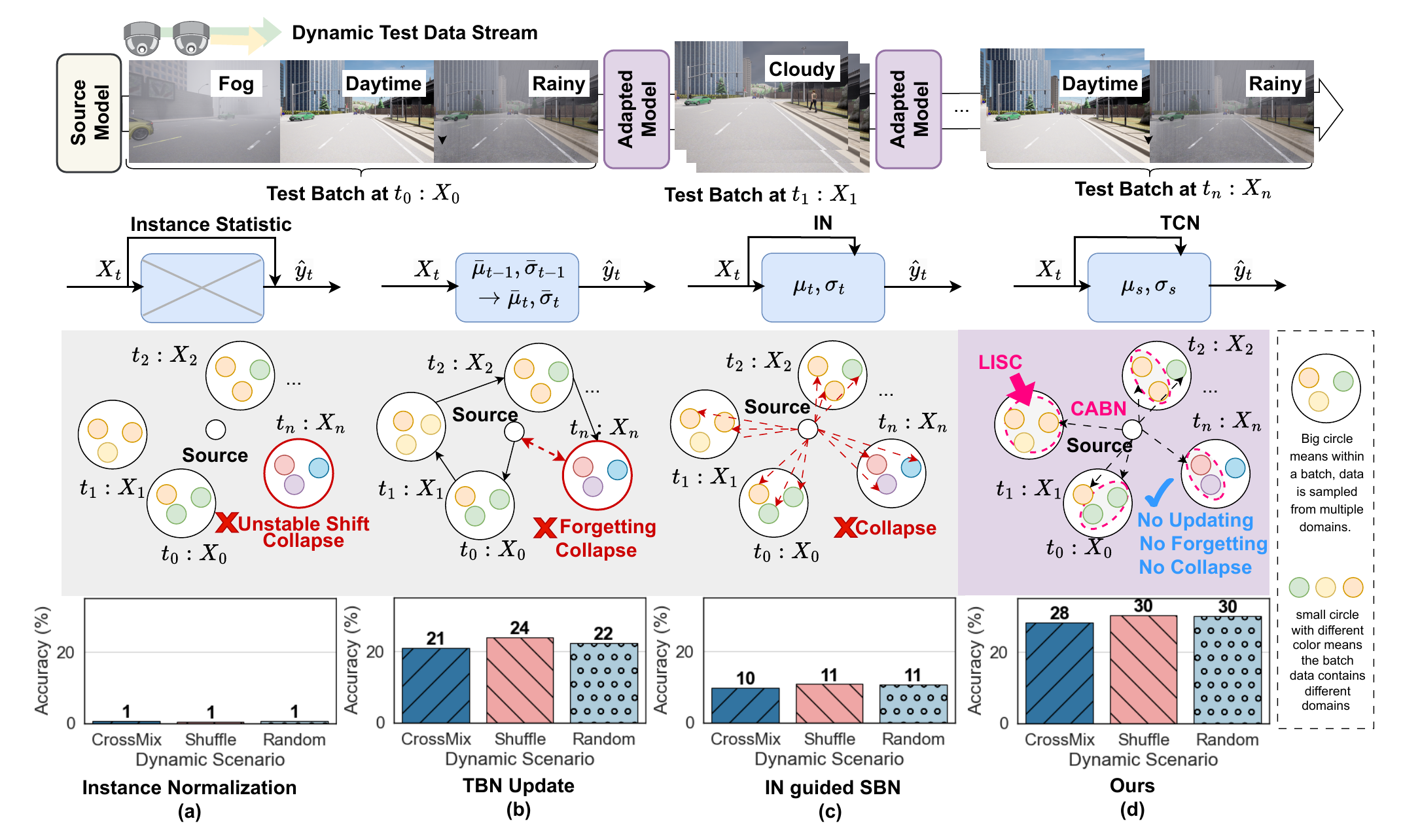}
                   \caption{
                   \textbf{Top}: illustration of test-time adaptation (TTA). \textbf{Bottom}: Comparison with different TTA methods. The proposed DYN can adjust the characterization of target domain feature distributions through LISC and CABN, enabling the model to maintain high performance and robustness under dynamic conditions}
    \label{fig:fig1}
\end{figure*}

Despite the considerable progress made with deep neural networks (DNNs), models trained on the source domain still suffer from a significant drop in performance when the test environment (e.g. target domains) differs significantly \cite{learningdataset,hendrycks2019benchmarking,choi2021robustnet,recht2019imagenet}. These changes in data distribution in multimedia applications (e.g. camera sensors, weather, and region) can lead to serious quality of experience (QoE) decline, even disastrous outcomes, particularly in risk-sensitive applications such as autonomous driving and traffic surveillance applications\cite{gongNOTERobustContinual2023}. To address this issue, \textit{test-time adaptation} (TTA) seeks to adapt models online without the source datasets and ground truth labels of test data streams \cite{wangTentFullyTesttime2021b} (See Figure~\ref{fig:fig1} Top).

Previous TTA studies typically involve two categories: \textit{Test-time fune-tuning}~\cite{wangTentFullyTesttime2021b,gongNOTERobustContinual2023,yuanRobustTestTimeAdaptation2023,liu2023vida,niuSTABLETESTTIMEADAPTATION2023,lee2024entropy,wang_decoupled_2024,gao_back_2023,chang_plug-and-play_2024} and \textit{Test-time Normalization} \cite{limTTNDomainShiftAware2023,mirza2022norm,wang2023dynamically,zhou_resilient_2024}. The former category focus on optimization of model parameters, such as through the partial backward: optimizing the affine parameters of models with self-supervision losses, such as entropy loss ~\cite{wangTentFullyTesttime2021b,gongNOTERobustContinual2023, yuanRobustTestTimeAdaptation2023,chen_contrastive_2022}, or in a fully backward manner: optimizing all parameters of models~\cite{wangContinualTestTimeDomain2022}. Although these methods can yield decent performance, they necessitate backward propagation and extra computational resources for acceleration, resulting in poor resource efficiency. Furthermore, the self-supervised training approach's dependence on the accuracy of pseudo-labels makes it susceptible to error accumulation and forgetting issues.

The second category focuses on Re-correcting batch normalization statistics using various batch normalization techniques. Instance normalization (IN)~\cite{pan2018two} (See Figure~\ref{fig:fig1}(a)) directly substitutes the source statistics with those from each instance, making it sensitive to target variations due to discarding the basic source knowledge, leading to instability. Test-Time Batch Normalization Update~(See Figure~\ref{fig:fig1}(b)) investigate the effects of updating historical statistics using test batch normalization statistics. In instance-aware batch normalization (IABN)~\cite{gongNOTERobustContinual2023} (See Figure~\ref{fig:fig1}(c)), the distribution of the source batch normalization (SBN)~\cite{ioffeBatchNormalizationAccelerating2015} is adjusted through IN. However, these methods also suffer from error accumulation caused by abnormal target distributions leading to suboptimal adaptation performance.

These TTA studies have mainly focused on static data patterns, where the test data stream exhibits only minor changes. Specifically, test samples within a batch are drawn from static distributions - either a single domain or continuous domain drift~\cite{wangTentFullyTesttime2021b, wangContinualTestTimeDomain2022,gongNOTERobustContinual2023,yuanRobustTestTimeAdaptation2023}. However, real-world test data often demonstrates atypical, dynamic patterns. Within a batch of data, the test samples lie in a realm of unforeseen changes from one or multiple different distributions (See Figure~\ref{fig:fig1} Top). As Figure~\ref{fig:fig1} Bottom shows, existing methods exhibit substantial performance degradation under such dynamic test conditions.

To explore the reasons behind the performance degradation, we conducted a series of measurements to re-understand batch normalization techniques from the perspective of \textit{Class-Related Features (CRF)} and \textit{Class-Irrelevant Features (CIF)} (See Section~\ref{sec:motivation}), which revealed that: 
\begin{itemize}
    \item Combining Source model Batch Normalization (SBN) and Test-time Batch Normalization (TBN) can effectively characterize the
complete target domain distribution in both static and dynamic scenarios. 

    \item The computation of TBN should ensure the class-irrelevant features it represents have high similarity.
    
\end{itemize}

Motivated by these observations, we propose a test-time normalization based approach (see Figure~\ref{fig:fig1} (d) ), called \textbf{D}iscover \textbf{Y}our \textbf{N}eighbours (DYN) for dynamic TTA scenarios. The main idea of DYN is by identifying samples with similar feature distributions within a batch and forming multiple clusters, we can replace the statistical values of TBN with those of test cluster normalization (TCN), thereby providing a consistent distribution of category-independent features. Specifically, our proposed DYN is implemented by injecting layer-wise instance statistics clustering (LISC) and cluster-aware batch normalization (CABN). Specifically, in LISC, we perform clustering on feature map with similar feature representations at each BN layer by calculating the cosine similarity of IN statistics. In CABN, we aggregate the statistical values of SBN and TCN to collaboratively characterize the feature distribution of the target domain, enabling the model to obtain a more robust representation. Our contributions can be summarized as follows:   

\begin{itemize}
\item \textbf{Novelty.} To our knowledge, this is the first backward-free TTA method addressing dynamic test patterns, and we conduct sufficient measurements to re-understanding of the batch normalization statistics through class-related and class-irrelevant features. 

\item \textbf{Effectiveness.} We propose a test-time normalization approach that utilizes instance normalization statistics to cluster samples with similar category-independent distributions. Combining TCN and SBN statistics enables robust representations adaptable to dynamic data.

\item \textbf{Promising Results.} Experiments on benchmark datasets demonstrate robust performance compared to state-of-the-art studies under dynamic distribution shifts, with up to a 35\% increase in accuracy.

\end{itemize}

\section{Motivation: Rethinking SBN and TBN}
\label{sec:motivation}

\begin{figure*}[htbp]
    \centering
    \begin{subfigure}[t]{0.23\textwidth}
        \centering
    \includegraphics[width=\linewidth,keepaspectratio]{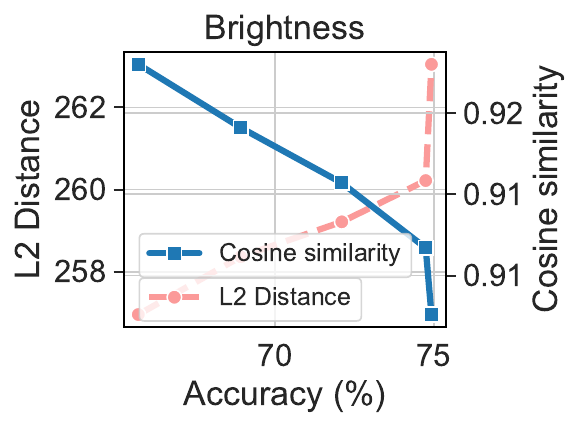} 
        \label{fig:motivation-level(a)}
    \end{subfigure}
    \begin{subfigure}[t]{0.23\textwidth}
        \centering
        \includegraphics[width=\linewidth,keepaspectratio]{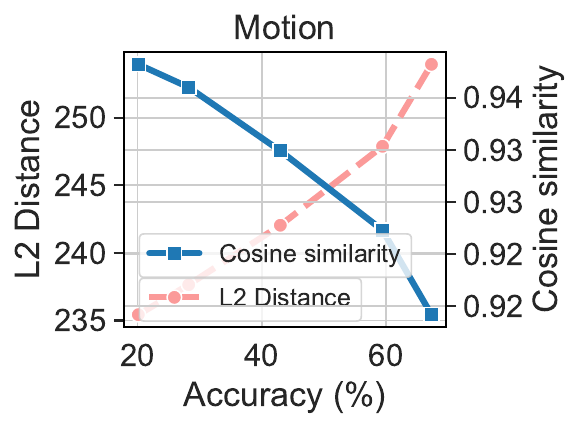} 
        \label{fig:motivation-level(b)}
    \end{subfigure}
    \begin{subfigure}[t]{0.23\textwidth}
        \centering
        \includegraphics[width=\linewidth,keepaspectratio]{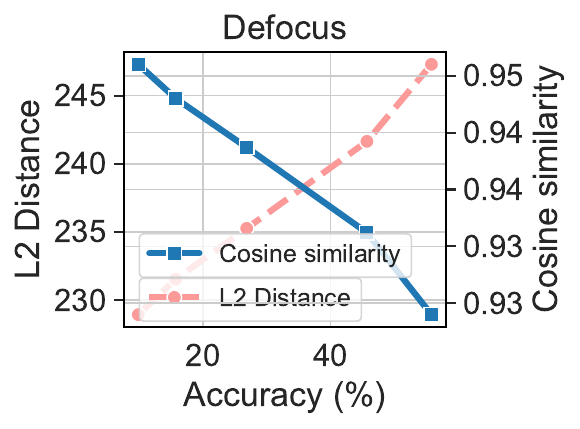}
        \label{fig:motivation-level(c)}
    \end{subfigure}
    \begin{subfigure}[t]{0.23\textwidth}
        \centering
        \includegraphics[width=\linewidth,keepaspectratio]{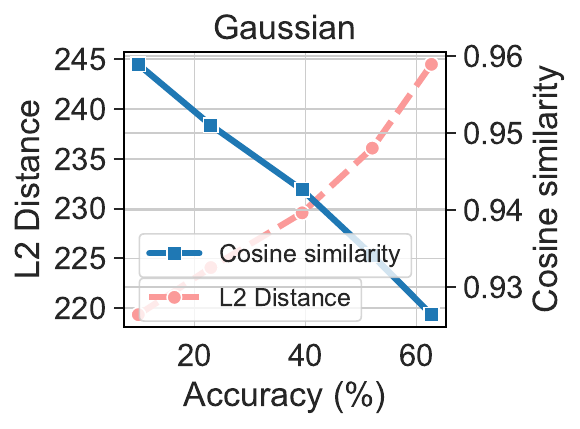}
        \label{fig:motivation-level(d)}
    \end{subfigure}
        \vspace{-0.2cm}
    \caption{TBN-IN distance vs. accuracy under different corruptions. The 5 data points in the figure represent samples with corruption levels 1 through 5, where higher levels correspond to lower accuracy. We compute the average distance between per-sample IN statistics and TBN statistics in the deep layers, reflecting the dispersion of feature distributions within a batch. It reveals that the distribution of CIF interferes with the distribution of CRF: as the sample corruption level increases, the feature distributions within a batch become more coupled.}
\label{fig:motivation-level}
    \vspace{-0.2cm}
\end{figure*}

\begin{figure*}[htbp]
    \centering
    \begin{subfigure}[!t]{0.22\textwidth}
        \centering
\includegraphics[width=\linewidth,keepaspectratio]{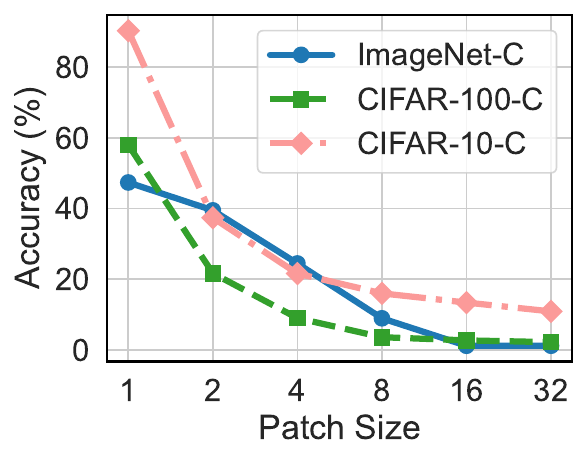}
    \caption{Patch size vs. accuracy }
    \label{motiviation-2:a}
    \end{subfigure}
    \begin{subfigure}[!t]{0.27\textwidth} 
    \includegraphics[width=\linewidth]{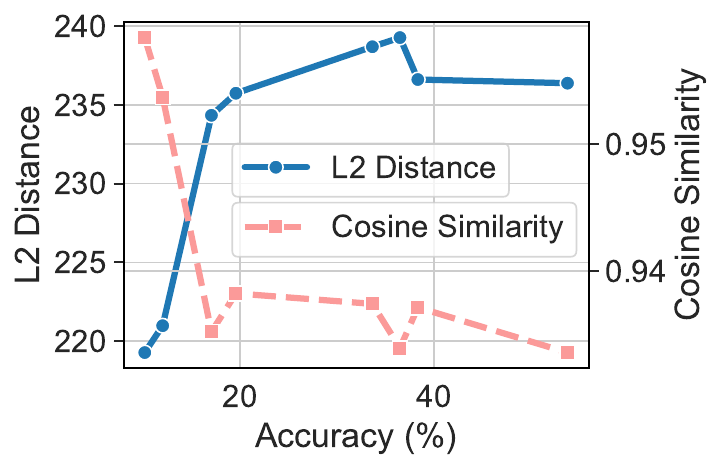} 
        \caption{TBN-IN distance vs. accuracy }
        \label{motiviation-2:b}
    \end{subfigure}
    \begin{subfigure}[!t]{0.24\textwidth} \includegraphics[width=\linewidth,keepaspectratio]{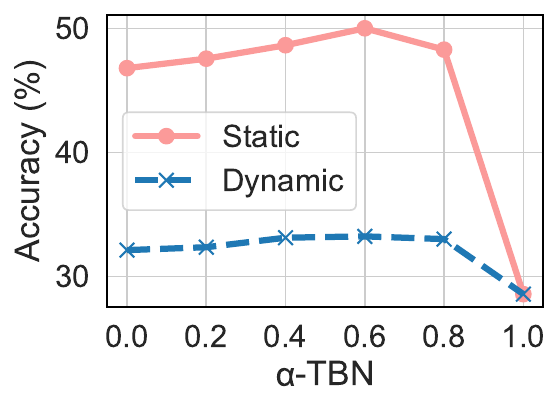} 
        \caption{$\alpha$ levels vs. accuracy}
        \label{motiviation-2:c}
    \end{subfigure}
    \begin{subfigure}[!t]{0.24\textwidth}
\includegraphics[width=\linewidth,keepaspectratio]{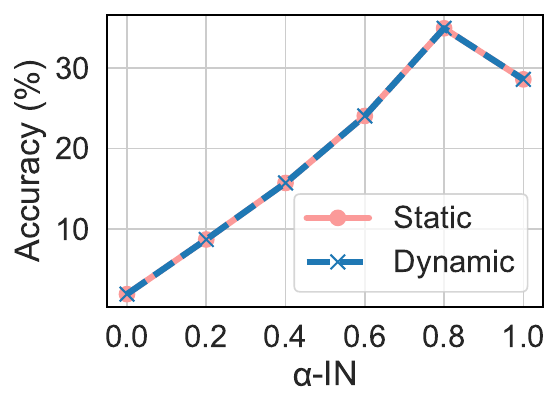}
        \caption{$\alpha$ levels vs. accuracy}
     \label{motiviation-2:d}
    \end{subfigure}
    \caption{(a) shows the inference accuracy of the pre-trained model after splitting images into varying numbers of patches and shuffling them. As the image is divided into smaller patches, feature reuse becomes more difficult for the model, resulting in lower inference accuracy. (b) displays the model's inference accuracy on samples with different corruptions versus the dispersion of per-sample IN statistics within a batch. Each data point represents samples from a distinct corruption type (level 5). (c) and (d) show the model's inference accuracy when combining TBN with SBN and when combining IN with SBN, respectively. In both figures, $\alpha$ denotes the proportion of SBN statistics used in the combination.}
\label{motiviation-2}
    \vspace{-0.2cm}
\end{figure*}

In this section, we will analyze the advantages and limitations of different types of normalization techniques from the perspective of class-related features and class-irrelevant features. Furthermore, we reveal the underlying reason for the ineffectiveness of all existing normalization methods in dynamic scenarios (experimental parameters are in the Supplementary Materials).

In deep neural networks, as the layers become deeper, there is a gradual occurrence of covariate shift internally. BN layers mitigate the impact of covariate shifts by normalizing the drifted features back to the same distribution. Therefore, BN layers play a crucial role in alleviating the effects of covariate shifts. The specific normalization formula is as follows: 
\begin{equation}
    B N\left(\mu_{c}, \sigma_{c}\right)=\gamma \cdot \frac{\left(F_{; c ;}-\mu_{c}\right)}{\sqrt{\left(\sigma_{c}\right)^{2}+\varepsilon}}+\beta,
\end{equation}
where $\mu_{c}$ and $\sigma_{c}$ are the statistical values of the BN layer, $\gamma$ and $\beta$ are the affine parameters of the BN layer, $F_{; c ;}$ is the input to the BN layer, and c is the number of channels.

Both the statistical values of SBN and TBN capture the distribution of data. The difference is that SBN characterizes the data distribution over the entire training set, while TBN characterizes the data distribution of the current batch of samples. Here, we denote the statistical values of SBN as $\mu ^{s} $ and $\sigma^{s} $, and the statistical values of TBN as $\mu ^{T}$ and $\sigma^{T}$. Since a sample itself consists of both class-irrelevant features  (e.g., background, weather) and {class-related features } (e.g., shape, structure)~\cite{lee2024entropy}. 

\textit{It reflects that the distribution of class-irrelevant features and the distribution of class-related features together characterize the complete data distribution.}

\begin{figure}[htbp]
    \centering
        \includegraphics[width=0.475\textwidth]{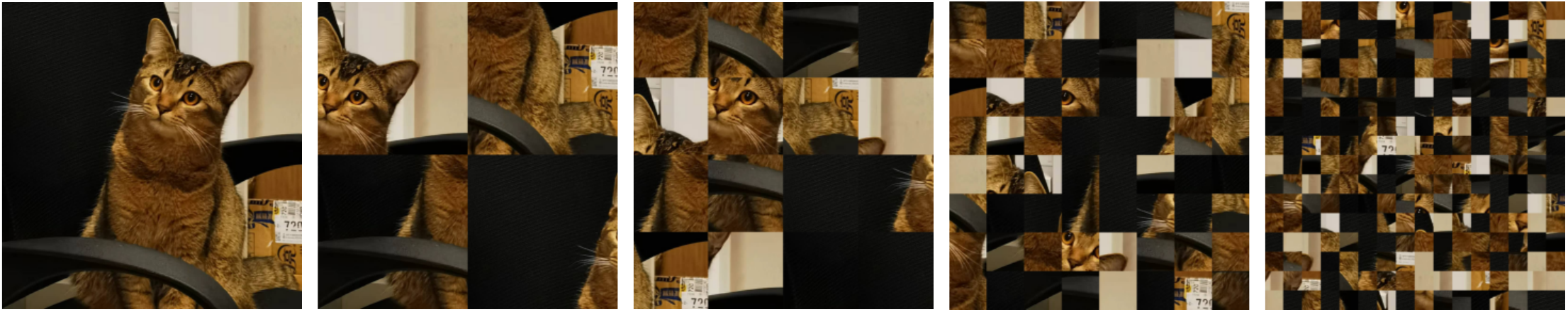} 
        \caption{Patch visualization. As the image is cropped into an increasing number of patches, the class-relevant features become harder to capture, while the class-irrelevant features are still retained. In this case, the inference accuracy of the model rapidly decline.}
        \label{fig:motivation-patch}
                \vspace{-0.2cm}
\end{figure}

\begin{figure}[htbp]
    \label{fig:motivation-bn}
    \centering
        \includegraphics[width=0.475\textwidth]{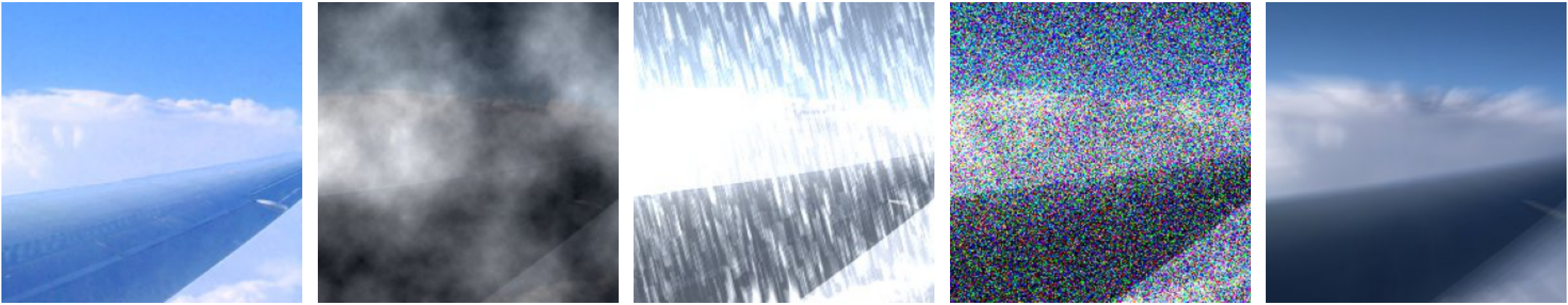} 
        \caption{Airplane wings in different corruption. More severe corruption perturbations of the image can obscure the class-relevant features, causing the overall feature distribution to become more entangled and thus decreasing the model's inference accuracy.}
                \vspace{-0.2cm}
\label{Fig_Corruption}
\end{figure}

\subsection{SBN: dominating CRF}
The model trained on the source domain is able to perform inference on the target domain dataset because the pre-trained model learns representations of class-related features while retaining the full class-related feature distributions in the SBN statistics from the training set. Therefore, when presented with target domain samples, the model can reuse the learned knowledge of these class-related features to inference, referred to as feature reuse~\cite{neyshabur2020being}.

As shown in Figure~(\ref{motiviation-2:a}), we evaluated the accuracy of the source model using different patch sizes. It reveals that the pre-trained model is still able to achieve reasonable accuracy when making inferences on the target domain. However, as we increasingly crop the target domain images into smaller patches and randomly shuffle them, the inference accuracy drops almost to 0. This occurs because the class-related features are completely destroyed by the patch structure, preventing the model from reusing this learned knowledge. Figure~\ref{fig:motivation-patch} is a visualization example of different patches.

Although directly using the pre-trained model and SBN statistics enables inference, the accuracy on the shifted target domain never reaches that on the source domain. \textit{This occurs because the class-irrelevant feature distributions change in the target domain, which cannot be captured solely by the SBN statistics.}

\subsection{TBN: the contributor of CIF}
$\mu^{T}$ and $\sigma^{T}$, computed over a batch of samples, capture distributions of both class-related and class-irrelevant features present in that batch. Therefore, previous methods such as Tent~\cite{wangTentFullyTesttime2021b} and BN adapt directly utilize TBN to achieve satisfactory results. However, relying solely on $\mu^{T}$ and $\sigma^{T}$ has two key limitations: 1) Insufficient or imbalanced batch samples can bias the estimated distribution of class-related features. 2) The distribution of class-irrelevant features in the batch can interfere with the modeled distribution of class-related features.

Figure~\ref{fig:motivation-level} shows model inference accuracy and distance between per-sample IN statistics and TBN statistics on batches containing ImageNet-C samples with corruption levels 1 to 5. Samples with level 1 corruption have the lowest domain shift. We calculate the L2 norm and cosine similarity distance between per-sample IN and TBN statistics in each BN layer. The average distance across samples indicates the coupling of feature distributions within a batch. As the corruption level increases, inference accuracy decreases and the distance between IN and TBN statistics declines, implying greater coupling of feature distributions in the batch. This demonstrates that a growing domain shift in class-irrelevant features can interfere with the TBN characterization of class-related features, reducing the differentiation of class-related feature distributions across categories. Figure~(\ref{motiviation-2:b})  presents consistent conclusions by testing different ImageNet-C corruption. Figure~\ref{Fig_Corruption} is a visualization example of Figure~(\ref{motiviation-2:b}).


\textit{Therefore, compared to SBN, TBN is more suitable as a contributor to class-irrelevant features, providing information about the distribution of class-irrelevant features in the target domain for SBN statistical values.}

\subsection{SBN with partial TBN: profiling the true feature distribution.} 
Figure ~(\ref{motiviation-2:c}) shows the inference performance of the model using SBN and TBN at different fusion ratios. Here, 'Static' represents samples within a batch that are independently and identically distributed, while 'Dynamic' indicates samples from multiple different distributions within a batch. Regardless of the scenario, combining the statistical values of SBN and TBN yields better results compared to using SBN or TBN statistical values alone. Moreover, when the proportion of SBN is higher, there is a greater improvement in the model's inference accuracy. This suggests that SBN's statistical values dominate the characterization of feature distribution by providing crucial information about class-related features, enabling the model to better reuse features. Meanwhile, TBN  provide information about the distribution of within the batch. Together, they capture the overall data distribution of the target domain.

However, regardless of the fusion ratio, the model's inference accuracy is significantly lower than in the 'Static' scenario. This reveals that when samples contain diverse class-irrelevant features from multiple distributions, the aggregation of these features in TBN results in high bias and inaccurate characterization of any distribution.

Figure~\ref{motiviation-2:d} shows the effect of fusing IN statistics, instead of TBN, with SBN values. Similar conclusions are reached in the 'Static' scenario, but performance differs for 'Dynamic'. Although IN is unaffected by class-irrelevant features of other samples, the high specificity of individual samples causes excessive self-normalization. Consequently, accuracy is not improved over TBN fusion and is even worse than 'Static'.  

The analysis yields two important takeaways: 

\begin{itemize}
    \item Combining SBN and TBN can effectively characterize the complete target domain distribution.
    \item The computation of TBN should ensure the class-irrelevant features it describes have high similarity.
\end{itemize}

Motivated by the above conclusions, we recognize that in dynamic scenarios, it is critical to prevent heterogeneous class-irrelevant features within TBN and differentiate samples in a batch. Grouping samples with similar individual normalized features enables reliable modeling of the class-irrelevant distribution. Based on these observations, we design our DYN methods described in the next section.

\section{Proposed Methods}
\subsection{Problem setup.} 
\subsubsection{TTA}
In test-time domain adaptation, we are given a model $f_\theta: \mathbf{x} \rightarrow y$ pre-trained on a source domain $\mathcal{D}_S$, models adapt to a target domain with $N$ number of test samples in the test set $\mathcal{D}_T$, $\left\{x_i,\right
\}_{i=1}^N \in \mathcal{D}_T$, without target labels provided. $\mathcal{D}_S$ and 
$\mathcal{D}_T$ share the output space. The goal of TTA is to adapt the model by utilizing continuously incoming test samples in an online fashion. For example, at each testing step $t$, the model $f_\theta$ receives a batch of instance $\mathbf{X}_t$ and simultaneously performs adaptation as well as produces predictions $\hat{Y}_t$. At the next step $t+1$, the model $f_\theta$ will perform adaptation and prediction on next batch of data $\mathbf{X}_{t+1}$ without access to previous data $\mathbf{X}_{t\rightarrow t}$.

\subsubsection{TTA in dynamic wild word}

In TTA, the target domain undergoes continuous changes over time, where samples inputted at the same moment maintain an independent and identically distributed (i.i.d.) property, i.e., $\mathbf{X}_{t}\in \mathcal{D}_t$ and $\mathcal{D}_t \ne \mathcal{D}_{t+1}$. However, as described in Section~\ref{sec:intro}, in certain specific scenarios, samples from different distributions may simultaneously aggregate into the model for inference. To simulate this practical scenario, we assume that at time step $t$, the target domain $\mathcal{D}_t$ is a set containing one or multiple domains, i.e., $\mathcal{D}_t = \left \{ \mathcal{D}_{t,1},{D}_{t,2},...,{D}_{t,N}\right \} $, where $N \ge  1$. This implies that $\mathbf{X}_{t} $ may originate from a single distribution or multiple different distributions.
\subsection{Layer-wise Instance Statistics Clustering}

The motivation analysis shows that when class-irrelevant features are similar across samples, using TBN provides superior performance to IN for modeling the class-irrelevant distribution. Clustering samples by feature similarity and computing TBN based on that cluster can mitigate excessive standardization from solely using IN. Moreover, as clustered samples share analogous features, it avoids high variance from individual samples, thus approximating TBN in the static scenario.  

Therefore, effective normalization requires reasonable partitioning of samples based on feature similarity. Clustering can accomplish this goal to some extent. However, most clustering algorithms rely on prior knowledge like predefined cluster centers or partition thresholds. Without reliable priors for novel domains, traditional clustering becomes ineffective. To address this, we adopt FINCH~\cite{sarfraz2019efficient}, an efficient, fully unsupervised, lightweight clustering algorithm. Hence, we propose a Layer-wise Instance Statistics Clustering (LISC) Algorithm. LISC partitions the layer-wise feature maps into distinct clusters based on instance statistics.

Specifically, given a feature map $ F \in \mathbb{R}^{B \times C \times H \times W}$ of a certain  BN layer, where $B$ represents the batch size, $C$ represents the number of channels, and $H$ and $W$ represent the height and width of the features, respectively. We need to group samples that have similar features based on the instance-wise statistical values $\mu^{I}$ for each sample, where $\mu^{I} \in \mathbb{R}^{C}$. The calculation of $\mu^I$ and the measure of sample feature similarity can be described as follows:
\begin{equation}
 \mu_{i, c}^{I}=\frac{1}{H \mathrm{W} } \sum_{l} f_{i ; c ; l},
\end{equation}
\begin{equation}
\operatorname{Cos}(i, j)=\frac{\mu_{i, c}^{I} \cdot \mu_{j, c}^{I}}{\left(\left\|\mu_{i, c}^{I}\right\| \left\|\mu_{j, c}^{I}\right\|\right)},
\label{Equa_dis}
\end{equation}
where $i$ and $j$ represent two different samples. Based on the cosine similarity of instance-wise statistical values for different samples, LISC clusters the samples at this BN layer.

FINCH discovers cluster structures in the data based on the first-neighbor relationships of samples. The algorithm assumes that the first neighbor of each sample is the most relevant and constructs clusters by progressively connecting neighbors. Compared to traditional clustering algorithms, It does not require setting distance thresholds, the number of clusters, or other hyperparameters, making it highly versatile and scalable. Given the integer indices of the first neighbors for each $\mu^{I}$, we compute an adjacency linking matrix:
\begin{equation}
    A(i, j)=\left\{\begin{array}{c}
1, \text { if } n_{i}^{1}=j \text { or } n_{j}^{1}=i \text { or } n_{i}^{1}=n_{j}^{1} \\
0, \text { otherwise }
\end{array}\right.,
\label{equa_A}
\end{equation}
where $n_{i}^{1}$ represents the first neighbor of the $i-th$ sample, which is the sample that minimizes $\operatorname{Cos}\left(i, n_{i}^{1}\right)$. The detailed algorithmic flow of the clustering process is provided in the Appendix.

\subsection{Cluster Aware Batch Normalization}
After clustering, the original feature map $F$ in the BN layer is divided into $k$ clusters, i.e., $F=\left\{C_{1}, C_{2}, \ldots ,C_{2},\ldots C_{k}\right\}$. $C_i \in \mathbb{R}^{b \times C \times H \times W}$, where $b$ represents the number of samples included in that cluster. Similar to computing BN-wise statistical values, we calculate the Test Cluster batch Normalization (TCN) statistical values $\mu_{c}^{i}$ and $\sigma_{c}^{i}$ for each cluster $C_{i}$ as follows:
\begin{equation}
   \mu_{c}^{i}=\frac{1}{ { b H W }} \sum_{b, l} f_{b ; c ;l},
\end{equation}
\begin{equation}
    \sigma_{c}^{i}=\sqrt{\frac{1}{b H W} \sum_{b, l}\left(f_{b ; c ; l}-\mu_{c}^{i}\right)^{2}}.
\end{equation}

As explained in the motivation, the SBN statistics $\mu_{s}$ and $\sigma_{s}$ can capture feature distributions and dispersion levels comprehensively for i.i.d. data. Aggregating the SBN cluster statistics with TCN is beneficial for feature reuse in pre-trained models. We thus propose Cluster Aware Batch Normalization (CABN), which utilizes SBN to correct the TCN statistics:
\begin{equation}
    \mu_{CABN}^{i}=\alpha \mu_{s}^{i}+(1-\alpha) \mu_{c},
\end{equation}
\begin{equation}
    \sigma_{CABN}^{i}=\alpha \sigma_{s}^{i}+(1-\alpha) \sigma_{c}.
\end{equation}

Here, $\alpha$ controls the mixing ratio. $\alpha=0$ uses only TCN while $\alpha=1$ uses only SBN. Mixing the TCN and SBN statistical values allows us to utilize the majority of knowledge learned in the pre-trained model for inference, while a small amount of TCN involvement corrects the noise introduced by the style and background features specific to that type of sample. Ultimately, the output of CABN is given by:
\begin{equation}
    \operatorname{CABN}_{i}\left(\mu_{CABN}^{i}, \sigma_{CABN}^{i}\right)=\gamma \cdot \frac{\left(f_{; c ;}-\mu_{CABN}^{i}\right)}{\sqrt{\left(\sigma_{CABN}^{i}\right)^{2}+\varepsilon}}+\beta, 1 \leq i \leq k ,
\end{equation}
where $\gamma$ and $\beta$ represent the affine parameters of the BN layer, and $\varepsilon$ is a small bias to prevent division by zero. This output indicates that TCN outputs personalized normalization results for each cluster, and there are differences between clusters.


\section{Experiments}
\definecolor{darkblue}{rgb}{0.0, 0.0, 0.55}
\begin{table*}[h]
    \caption{Comparisons with state-of-the-art methods on CIFAR10-C, CIFAR100-C and ImageNet-C respectively (severity level = 5) under \textbf{\textsc{Batch Size=64}} regarding \textbf{Accuracy (\%)}. Each method was evaluated under the CrossMix, Random, and Shuffle scenarios using a ResNet-50 model architecture. The best result  is denoted in bold black font.
    }
    \label{tab:imagenet-c-bs1-level3}
\newcommand{\tabincell}[2]{\begin{tabular}{@{}#1@{}}#2\end{tabular}}
 \begin{center}
 \begin{threeparttable}
 \Huge
    \resizebox{1.0\linewidth}{!}{
  \begin{tabular}{l|ccc>{\columncolor{blue!8}}c|ccc>{\columncolor{blue!8}}c|ccc>{\columncolor{blue!8}}c|>{\columncolor{darkblue!8}}c}
  
  \multicolumn{1}{c}{} & \multicolumn{4}{c}{CrossMix} & \multicolumn{4}{c}{Random} & \multicolumn{4}{c}{Shuffle}  \\

  Method & CIFAR10-C & CIFAR100-C& ImageNet-C & Avg.& CIFAR10-C & CIFAR100-C& ImageNet-C & Avg. & CIFAR10-C & CIFAR100-C& ImageNet-C & Avg.& Avg-All\\

  \midrule
  Source (ResNet-50) & 57.39  & 28.59  & 25.64  & 37.21  & 57.38  & 28.58  & 25.93  & 37.30  & 57.38  & 28.58  & 25.80  & 37.25 & 37.25 \\ 
   \rowcolor[HTML]{d0d0d0}
         \multicolumn{14}{c}{\Huge\textbf{\textsc{Test-Time Fine-Tune}}} \\
         ~~$\bullet~$EATA & 61.97 & 32.74 & 19.68 & 38.13 & 68.25 & 42.15 & 24.26 & 44.89 & 67.83 & 41.37 & 24.50 & 44.57 &  42.53\\
        ~~$\bullet~$DeYO & 68.85 & 30.43 & 19.13 & 39.47 & \textbf{75.63} & 36.67 & 24.32 & 45.54 & \textbf{75.62} & 35.45 & 25.21 & 45.43 &  43.48 \\ %
        ~~$\bullet~$SAR    & 61.70 & 31.45 & 18.65 & 37.27 & 71.04 & 40.92 & 23.62 & 45.19 & 71.49 & 39.58 & 23.50 & 44.86 & 42.44 \\

        ~~$\bullet~$TENT   & 62.77 & 31.57 & 18.45 & 37.60 & 71.50 & 40.96 & 23.15 & 45.20 & 71.56 & 40.73 & 23.78 & 45.36 & 42.72 \\ 
        ~~$\bullet~$NOTE & 63.03 & 32.96 & 17.44 & 37.81 & 62.81 & 33.19 & 21.64 & 39.21 & 65.34 & 35.12 & 22.98 & 41.15 & 39.39 \\ 
        ~~$\bullet~$ViDA  & 61.97 & 32.14 & 18.52 & 37.54 & 67.96 & 39.48 & 23.33 & 43.59 & 67.96 & 39.48 & 23.06 & 43.50 & 41.54 \\\ 
     
         ~~$\bullet~$RoTTA & 43.70 & 24.05 & 21.85 & 29.87 & 48.68 & 23.80 & 20.39 & 30.96 & 54.79 & 29.29 & 22.71 & 35.60 & 32.14 \\

 \rowcolor[HTML]{d0d0d0}
         \multicolumn{14}{c}{\Huge\textbf{\textsc{Test-Time Normalization}}} \\

       ~~$\bullet~$TBN & 61.96 & 32.12 & 18.72 & 37.60 & 67.75 & 39.16 & 22.99 & 43.30 & 67.63 & 39.22 & 23.35 & 43.40 & 41.43 \\ 
       ~~$\bullet~$$\alpha$-BN   & 62.41 & 33.22 & 21.92 & 39.18 & 69.88 & 41.59 & 27.57 & 46.35 & 69.87 & 41.60 & 27.78 & 46.42 & 43.98 \\

       ~~$\bullet~$IABN    & 62.63 & 24.54 & 9.73  & 32.30 & 64.59 & 26.40 & 10.75 & 33.91 & 64.59 & 26.40 & 10.79 & 33.93 &  33.38 \\
      
       \rowcolor{pink!30}~~$\bullet~$DYN (ours)  &\textbf{ 71.54}&  \textbf{ 39.75 }& \textbf{28.17} & \textbf{ 46.49}  & 73.09 & \textbf{ 42.56 } & \textbf{ 30.12 } & \textbf{48.59} & 72.74  & \textbf{ 42.87} & \textbf{ 30.00 } &\textbf{ 48.54 }  &\textbf{ 47.87 } \\

 \end{tabular}
 }
  \end{threeparttable}
  \end{center}
  \label{Table_All}
\end{table*}

\begin{figure}[b]{}
        \label{fig:DifferntCluster}
    \includegraphics[width=\linewidth,keepaspectratio]{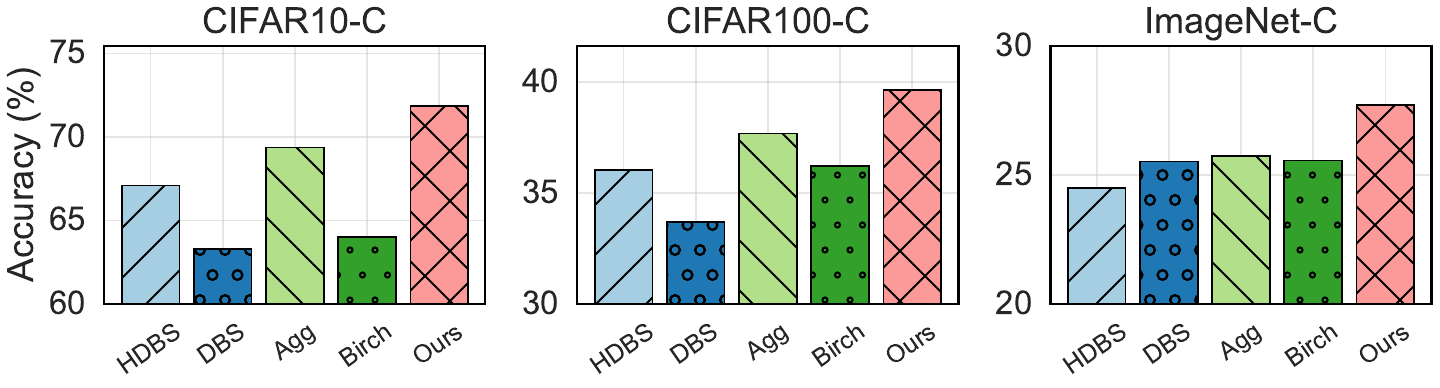}
    \caption{Different cluster methods vs. accuracy. The proposed LISC of us can effectively aggregate similar class-irrelevant features, thereby improving the model's inference performance.}
\label{Fig_Cluster}
\end{figure}

\subsection{Experimental Setup}

We achieved the proposed method DYN and baselines on the TTAB framework \cite{zhao2023ttab}. Additional experimental details, e.g., hyperparameters of the baselines and the dataset selections, are specified in the Appendix.

\textbf{{Environment and Hyperparameter Configuration.} }
 All the experiments in this article were conducted on the NVIDIA GeForce RTX 4090 GPU. The experimental code is implemented based on PyTorch 1.10.1 and Python 3.9.7. The aggregation parameter $\alpha$ in CABN is set to 0.8.

\textbf{{Baselines.}}
 We consider the following baselines, including state-of-the-art test-time adaptation and test-time normalization algorithms:
\begin{itemize}

\item \textbf{Test-time fine-tune.} The \textbf{Source} \textbf{(Source)} evaluates the model trained on the source data directly on the target data without any adaptation. Sharpness-aware and reliable entropy minimization method \textbf{(SAR)} \cite{niuSTABLETESTTIMEADAPTATION2023} has the advantage of conducting selective entropy minimization, excluding samples with noisy gradients during online adaptation, which leads to more robust model updates. Additionally, SAR optimizes both entropy and the sharpness of the entropy surface simultaneously, ensuring the model's robustness to samples with remaining noisy gradients.  Efficient anti-forgetting test-time adaptation \textbf{(EATA)} \cite{niu2022efficient} improves the stability of model updates by filtering high-entropy samples, while applying Fisher regularizer to limit the extent of changes in important parameters, thereby alleviating catastrophic forgetting after long-term model adaptation. Destroy your object \textbf{(DeYo)} \cite{lee2024entropy} disrupts the structural class-related features of samples by chunking them, and selects appropriate samples for model adaptation by comparing the entropy change in predictions before and after chunking, thereby enabling the model to learn the correct knowledge.  Test entropy minimization \textbf{(TENT)} \cite{wangTentFullyTesttime2021b} optimizes the model for confidence as measured by the entropy of its predictions and estimates normalization statistics and optimizes channel-wise affine transformations to update online on each batch. Non-i.i.d. Test-time adaptation \textbf{(NOTE)} \cite{gongNOTERobustContinual2023} is mainly two-fold: Instance-Aware Batch Normalization (IABN) that corrects normalization for out-of-distribution samples, and Prediction-balanced Reservoir Sampling (PBRS) that simulates i.i.d. data stream from non-i.i.d. stream in a class-balanced manner. Robust test-time Adaptation \textbf{(RoTTA)} \cite{yuanRobustTestTimeAdaptation2023}  share a similar approach with note, which simulates an i.i.d. data stream by creating a sampling pool and adjusting the statistics of the batch normalization (BN) layer. Visual Domain Adapter \textbf{(ViDA)} \cite{liu2023vida} shares knowledge by partitioning high-rank and low-rank features. For the aforementioned methods that require updating the model, we follow the online-TTA setup. We assume that the source data, which is used for model pre-training, is not available for use in test-time adaptation (TTA). We conduct online adaptation and evaluation, continuously updating the model.

\item \textbf{Test-time normalization.} \textbf{TBN}  uses the mean and variance of the current input batch samples as the statistics for the BN (Batch Normalization) layer. \textbf{$\alpha$-BN}  aggregates the statistics of TBN and SBN (Spatial Batch Normalization) to obtain new statistics for the BN layer. \textbf{IABN} is a method for calculating BN layer statistics in NOTE, which involves using the statistics of IN (Instance Normalization) of the samples for correction. For these backward-free methods, we also follow the online-TTA setting. Additionally, we do not make any adjustments to the model parameters. Instead, we only modify the statistics of the BN layer during the inference process using different approaches.
    
\end{itemize}

Our general setting for the baseline result \cite{zhao2023ttab,gongNOTERobustContinual2023} involves using a test batch size of 64 and performing one adaptation epoch for adaptation.  Additionally, we choose the method-specific values for the hyperparameters reported in their papers or the official codes \cite{zhao2023ttab}.

\textbf{Datasets.}
We utilize the CIFAR10-C, CIFAR100-C, and ImageNet-C datasets \cite{hendrycks2019benchmarking} employed in the TTA benchmark \textbf{(TTAB)} \cite{zhao2023ttab} to assess the models' robustness to corruptions.  Each type of corruption has 5 severity levels, where a higher level indicates a greater degree of data drift. CIFAR10-C and CIFAR100-C are small-scale datasets, each comprising 10 and 100 classes, with 50,000 training samples and 10,000 test samples. In contrast, ImageNet-C is a large-scale dataset with 1,000 classes, consisting of 1,281,167 training samples and 50,000 test samples. 
Similar to previous studies \cite{zhao2023ttab}, we set the corruption level to the most severe level of 5. We utilize ResNet50 \cite{heDeepResidualLearning2015} as the backbone network and pre-train the ResNet50 models using clean data such as CIFAR10/CIFAR100/ImageNet. 


\textbf{{Scenarios.}} In our experiments, we employed three scenarios: CrossMix, Random, and Shuffle. CrossMix indicates that each batch of input samples is drawn from multiple different distributions. Random indicates that the input samples in each batch are randomly switched between multiple different distributions or maintained as i.i.d. Shuffle indicates that the input samples in each batch remain i.i.d., but batches containing samples from different distributions are mixed in intermittently. 

\subsection{Experimental Results and Analysis under Different Scenario Settings}


Here we conducted experiments in three different scenarios: CrossMix, Random and Suffle. Following the experimental setup of previous studies, we selected the most severely corrupted sample (level 5) from each corruption type.

Table~\ref{Table_All} shows the performance of different TTA methods. From the table, it can be observed that our method significantly outperforms other baselines in terms of average accuracy across the three scenarios. Particularly, our method exhibits the greatest advantage in the CrossMix scenario, with an average accuracy approximately 17\% higher than the lowest baseline (RoTTA) and still 7\% higher than the highest baseline (DeYO). This indicates that our method possesses inherent advantages when dealing with batch data containing multiple distributions. By replacing TBN with CABN, our method effectively captures class-agnostic feature distributions for each cluster. In the remaining two scenarios, our method maintains robustness. In the Random scenario, our average accuracy surpasses the lowest baseline (RoTTA) by approximately 18\% and exceeds the highest baseline ($\alpha$-BN) by around 3\%. In the Shuffle scenario, our average accuracy outperforms the lowest baseline (RoTTA) by approximately 18\% and surpasses the highest baseline ($\alpha$-BN) by around 13\%. This demonstrates the effectiveness of our method in handling dynamically changing distribution patterns. Whether the batch data originates from the same distribution, different distributions, or undergoes real-time changes between the two, our method consistently achieves good performance.

\subsection{Performance Comparison Between LISC (Ours) and Other Clustering Algorithms}
Figure~\ref{Fig_Cluster} displays the performance of LISC and other clustering algorithms on CIFAR10-C, CIFAR100-C and ImageNet-C respectively. Among these, HDBS \cite{mcinnes2017hdbscan} and DBS \cite{sander1998density} are density-based algorithms, while Agg \cite{murtagh2014ward} and BIRCH \cite{zhang1996birch} are hierarchy-based. With our LISC method, the model inference accuracy surpasses other clustering approaches on all three datasets. This demonstrates that LISC can effectively aggregate class-irrelevant features with high similarity, thereby providing a cleaner representation of the class-irrelevant distribution

\subsection{Performance under Different Model Structures}
\begin{table}[htbp]
    \caption{Comparisons with state-of-the-art methods on CIFAR10-C and CIFAR100-C respectively (severity level = 5) under \textbf{\textsc{Batch Size=64}} regarding \textbf{Accuracy (\%)}. Each method was evaluated under the CrossMix scenario with   using a ResNet-26 model architecture. The best result  is denoted in bold black font.
    }
    \label{tab:imagenet-c-bs1-level3}
\newcommand{\tabincell}[2]{\begin{tabular}{@{}#1@{}}#2\end{tabular}}
 \begin{center}
 \begin{threeparttable}
  \scriptsize
    \resizebox{0.8\linewidth}{!}{
  \begin{tabular}{l|cc>{\columncolor{blue!8}}c}
  

  Method & CIFAR10-C & CIFAR100-C & Avg.\\

  \midrule
  Source (ResNet-26)& 53.06 & 31.34 &  42.20 \\
   \rowcolor[HTML]{d0d0d0}
         \multicolumn{4}{c}{\scriptsize\textbf{\textsc{Test-Time Fine-Tune}}} \\
         ~~$\bullet~$EATA & 59.34 & 34.24 & 46.79  \\ 
        ~~$\bullet~$DeYO & 65.47 & 33.89 &  49.68  \\
        ~~$\bullet~$SAR    & 58.70 & 30.42 & 44.56  \\

        ~~$\bullet~$TENT   & 59.01 & 30.37 & 44.69   \\ 
        ~~$\bullet~$NOTE & 60.18 & 31.28 & 45.73    \\ 
        ~~$\bullet~$ViDA & 65.80 & 30.52 &  48.16 \\ 
       
         ~~$\bullet~$RoTTA & 49.26 & 20.10 &  34.68   \\

 \rowcolor[HTML]{d0d0d0}
         \multicolumn{4}{c}{\scriptsize\textbf{\textsc{Test-Time Normalization}}} \\

       ~~$\bullet~$TBN & 59.34 & 30.52 &  44.93   \\
       ~~$\bullet~$$\alpha$-BN   & 58.51 & 32.81 &  45.66  \\ 
       ~~$\bullet~$IABN    & 62.31	 &19.72  &  41.02   \\
      
       \rowcolor{pink!30}~~$\bullet~$DYN (ours)  & \textbf{68.21} & \textbf{38.24} & \textbf{53.22}  \\
        
 \end{tabular}
 }
  \end{threeparttable}
  \end{center}
  \label{Table_Res26}
\end{table}

Table~\ref{Table_Res26} shows the effects of the method proposed in this paper on the pre-trained model with different structure (ResNet-26). The experiment was conducted in the CrossMix scenario. From the table, it can be seen that even when applied to different model structures, our method still outperforms all other baselines. On CIFAR10-C, our method achieves an accuracy approximately 3\% higher than the best-performing baseline (Vida). On CIFAR100-C, our method achieves an accuracy approximately 4\% higher than the best-performing baseline (EATA). This indicates that our method performs well across different model structures and exhibits high robustness to model variations
\begin{table*}[htbp]
    \caption{Comparisons with state-of-the-art methods on CIFAR10-C, CIFAR100-C and ImageNet-C respectively (severity level = 5) under \textbf{\textsc{Batch Size=64}} regarding \textbf{Accuracy (\%)}. Each method was evaluated under the CrossMix scenario with  various numbers of domains using a ResNet-50 model architecture. The best result  is denoted in bold black font.
    }
    \label{tab:imagenet-c-bs1-level3}
\newcommand{\tabincell}[2]{\begin{tabular}{@{}#1@{}}#2\end{tabular}}
 \begin{center}
 \begin{threeparttable}
 \Huge
    \resizebox{1.0\linewidth}{!}{
  \begin{tabular}{l|ccc>{\columncolor{blue!8}}c|ccc>{\columncolor{blue!8}}c|ccc>{\columncolor{blue!8}}c|>{\columncolor{darkblue!8}}c}
  
  \multicolumn{1}{c}{} & \multicolumn{4}{c}{CIFAR10-C} & \multicolumn{4}{c}{CIFAR100-C} & \multicolumn{4}{c}{ImageNet-C}  \\

  Method & 6 Domains & 9 Domains& 12 Domains & Avg.& 6 Domains &9 Domains& 12 Domains & Avg. & 6 Domains & 9 Domains& 12 Domains & Avg.& Avg-All\\

  \midrule
  Source (ResNet-50)& 46.59 & 52.54 & 55.84 & 51.66 & 19.23 & 24.25 & 27.15 & 23.54 & 16.21 & 20.07 & 25.20 & 20.49 & 31.90 \\
   \rowcolor[HTML]{d0d0d0}
         \multicolumn{14}{c}{\Huge\textbf{\textsc{Test-Time Fine-Tune}}} \\
         ~~$\bullet~$EATA & 57.67 & 60.89 & 63.79 & 60.78 & \textbf{35.52} & 35.05 & 33.34 & 34.64 & 6.97  & 13.24 & 18.47 & 12.89 & 36.10 \\ 
        ~~$\bullet~$DeYO  & 65.23 & 69.22 & 69.89 & 68.11 & 35.35 & 35.71 & 33.96 & 35.01 & 7.39  & 13.59 & 17.85 & 12.94 & 38.69 \\
        ~~$\bullet~$SAR   & 59.76 & 63.76 & 65.75 & 63.09 & 31.03 & 33.48 & 33.71 & 32.74 & 7.44  & 12.75 & 18.05 & 12.75 & 36.19 \\

        ~~$\bullet~$TENT   & 60.79 & 64.69 & 66.59 & 64.02 & 30.33 & 33.07 & 33.44 & 32.28 & 7.41  & 12.69 & 17.24 & 12.45 & 36.25 \\ 
        ~~$\bullet~$NOTE & 58.61 & 61.88 & 64.93 & 61.81 & 28.55 & 31.46 & 33.65 & 31.22 & 12.92 & 17.23 & 22.71 & 17.62 & 36.88 \\ 
        ~~$\bullet~$ViDA & 57.64 & 60.89 & 63.77 & 60.77 & 27.89 & 31.04 & 32.96 & 30.63 & 6.80  & 12.09 & 18.03 & 12.31 &   34.57\\ 
       
         ~~$\bullet~$RoTTA & 52.03 & 35.44 & 42.26 &  43.24 & 23.44 & 21.73 & 23.72 & 22.96 & 6.96  & 16.16 & 22.37 & 15.16 &   27.12\\

 \rowcolor[HTML]{d0d0d0}
         \multicolumn{14}{c}{\Huge\textbf{\textsc{Test-Time Normalization}}} \\

       ~~$\bullet~$TBN & 57.63 & 60.89 & 63.77 &  60.76 & 27.89 & 31.03 & 32.95 & 30.62 & 7.39  & 12.39 & 17.51 & 12.43 & 34.60 \\
       ~~$\bullet~$$\alpha$-BN  & 58.17 & 61.68 & 64.27 & 61.37 & 28.49 & 31.79 & 33.81 & 31.36 & 9.78  & 15.36 & 21.80 & 15.65 & 36.13 \\

       ~~$\bullet~$IABN      & 55.30 & 59.41 & 64.24 & 59.65 & 18.99 & 21.99 & 25.62 & 22.20 & 3.55  & 6.44  & 10.30 &  6.76 &  29.54\\
      
       \rowcolor{pink!30}~~$\bullet~$DYN (ours)  & \textbf{66.59} & \textbf{70.30} & \textbf{72.97} & \textbf{69.95} & 35.08 & \textbf{38.82} & \textbf{40.68} & \textbf{38.19} & \textbf{17.05} & \textbf{22.38} & \textbf{27.40} &  \textbf{22.28} & \textbf{43.47} \\

 \end{tabular}
 }
  \end{threeparttable}
  \end{center}
  \label{Table_Domain}
\end{table*}

\subsection{Performance under Different Domain Scales}


Table~\ref{Table_Domain} shows the performance of the method proposed in this paper under different domain scales in CrossMix Scenario. It can be observed that our proposed approach consistently outperforms all other baselines as the number of class-agnostic feature distributions within the batch data increases from fewer to more (i.e., the composition of batch data transitions from simple to complex). In terms of overall average accuracy, our method surpasses the worst-performing baseline (RoTTA) by approximately 16\% and outperforms the best-performing baseline (DeYO) by around 5\%. Across the three different datasets, our method achieves an average accuracy that is approximately 2\%-5\% higher than the best-performing baseline. This demonstrates the excellent performance of our method in scenarios where the number of distribution variations within batch data dynamically changes.

\subsection{Performance under Different Batch Size}

\begin{figure*}[htbp]
    \centering
    \begin{subfigure}[t]{0.26\textwidth}
        \includegraphics[width=\textwidth,keepaspectratio]{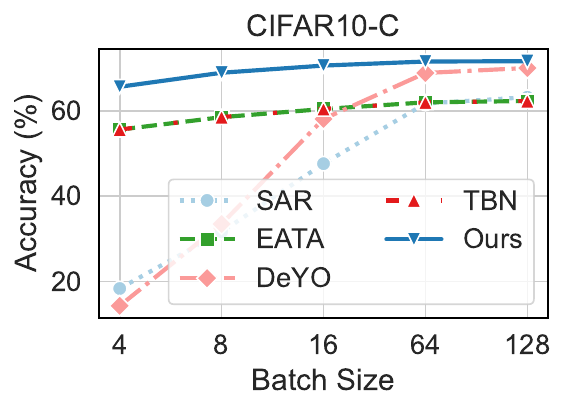} 
        \label{fig:batch-cifar10}
    \end{subfigure}
    \begin{subfigure}[t]{0.26\textwidth}
        \includegraphics[width=\textwidth,keepaspectratio]{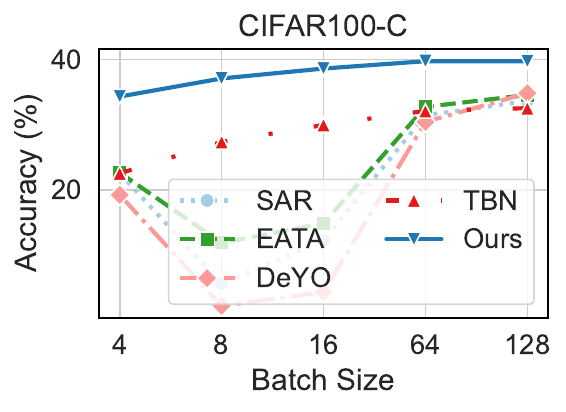} 
                    \label{fig:batch-cifar10}
    \end{subfigure}
    \begin{subfigure}[t]{0.26\textwidth}
        \includegraphics[width=\textwidth,keepaspectratio]{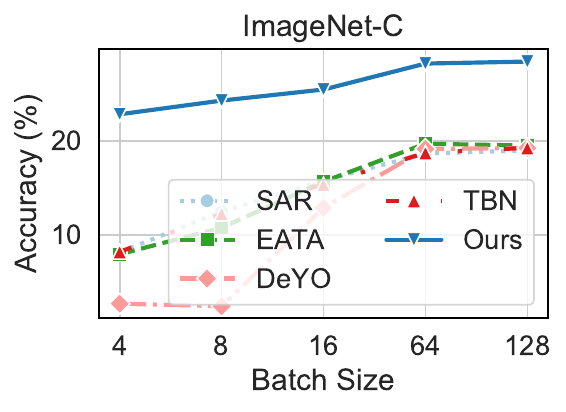}
                    \label{fig:batch-imagenet-c}
    \end{subfigure}
    \caption{Batch size vs. accuracy. Other methods exhibit poorer performance at low batch sizes, only stabilizing as batch size increases. In contrast, our approach is insensitive to batch size variation, demonstrating greater robustness.}
    \label{fig:motivation}
\label{Fig_Batch}
    \vspace{-0.4cm}
\end{figure*}

Figure~\ref{Fig_Batch} shows the performance of the method proposed in this paper under different batch size in CrossMix Scenario. From the table, it can be observed that our method exhibits stable performance across both small and large batch sizes for the three datasets, with only slight variations. In contrast, the other methods experience significant fluctuations. On CIFAR100-C and ImageNet-C, the accuracy of the other methods drops sharply as the batch size decreases, and it only stabilizes when the batch size exceeds 64. This indicates that our method has high robustness to batch size variations and can achieve good performance regardless of the batch size.

\section{Related Work}






Regarding online TTA in dynamic environments, existing research mainly focuses on online TTA, continual TTA, TTA in the open wild world.

\subsection{Online TTA}
In some literature, TTA is also referred to as unsupervised domain adaptation. Unlike domain adaptation, which requires access to both source and target data, TTA methods can adapt without any source domain data. Since test samples themselves contain informatio  showed that simply adjusting batch normalization statistics during testing can significantly improve model performance on corrupted data. Recent test-time adaptation methods further perform backpropagation to update the neural network weights of the model to adapt to the new data distribution. For instance, Wang et al. \cite{wangTentFullyTesttime2021b} utilized an entropy minimization strategy to update the affine parameters of the batch normalization layer. Zhang et al. \cite{zhang2022memo} employed entropy minimization to update all model parameters and artificially increased the batch size using test-time augmentation techniques to enhance the model's inference accuracy.

\subsection{Continual TTA}
TTA methods are typically designed to adapt the model to a single target domain. However, in the real world, deployed models are likely to encounter a series of domain shifts. Therefore, continual TTA considers the scenario of adapting online to continuous changes in the target domain. Some existing TTA methods can also be applied to continual adaptation, such as TENT proposed by Wang et al. \cite{wangTentFullyTesttime2021b}. However, solely relying on self-training methods can lead to error accumulation during continual adaptation, as it relies on pseudo-labels obtained from incorrect predictions. COTTA \cite{wangContinualTestTimeDomain2022} is the first dedicated method proposed for continual test-time adaptation. It mitigates the problem of error accumulation by using weighted averaging and augmented averaging predictions, and further incorporates random restoration to prevent catastrophic forgetting. EATA \cite{niuEfficientTestTimeModel} improves the reliability of pseudo-labels by filtering high-entropy samples and applies Fisher regularization to prevent excessive changes in model parameters.

\subsection{TTA in Dynamic Wild World}

In the real world, test data is often more complex, with drifted data that is difficult to maintain independently and identically distributed. It also encounters scenarios of mixed domains or imbalanced sample distributions. To address this, SAR \cite{niu2023towards} elucidates the main characteristics of wild data and analyzes the gradient features and underlying reasons for the performance degradation of domain-adaptive models when tested on wild data through experiments. To tackle the issue of non-independent and identically distributed labels in open compound scenarios, NOTE \cite{gongNOTERobustContinual2023} proposes the IN-based IABN method, which corrects the BN layer's statistical values and constructs a sample library with balanced labels through sampling. DeYo \cite{lee2024entropy} analyzes the proper learning approach for models from the perspectives of category-related features and category-independent features, and utilizes feature decoupling to enable the model to learn more category-related features.

While the aforementioned research provides some analysis on TTA in open environments, there is currently no dedicated method specifically addressing TTA in dynamic environments. To address this gap, we, for the first time, analyze the data distribution in dynamic environments from the perspective of correcting BN layer statistical values and propose a backward free solution.
\section{Conclusion}

This paper explores test-time adaptation for dynamic scenarios. We innovatively analyze source batch normalization (SBN) and test-time batch normalization (TBN) from the perspective of class-relevant and class-irrelevant feature distributions: SBN primarily captures class-irrelevant feature distributions while TBN is more suitable as a contributor for class-irrelevant feature distributions. Based on this, we propose a test-time normalization method called Discover Your Neighbors (DYN) tailored for dynamic settings. DYN identifying similar feature maps via instance normalization statistics and clustering into groups which provides consistent class-irrelevant representations. 
Experimental results validate DYN's robustness and effectiveness, demonstrating maintained performance under changing data stream patterns.
\clearpage
\bibliographystyle{ACM-Reference-Format}
\bibliography{sample-base,main}


\begin{thebibliography}{33}


\ifx \showCODEN    \undefined \def \showCODEN     #1{\unskip}     \fi
\ifx \showDOI      \undefined \def \showDOI       #1{#1}\fi
\ifx \showISBNx    \undefined \def \showISBNx     #1{\unskip}     \fi
\ifx \showISBNxiii \undefined \def \showISBNxiii  #1{\unskip}     \fi
\ifx \showISSN     \undefined \def \showISSN      #1{\unskip}     \fi
\ifx \showLCCN     \undefined \def \showLCCN      #1{\unskip}     \fi
\ifx \shownote     \undefined \def \shownote      #1{#1}          \fi
\ifx \showarticletitle \undefined \def \showarticletitle #1{#1}   \fi
\ifx \showURL      \undefined \def \showURL       {\relax}        \fi
\providecommand\bibfield[2]{#2}
\providecommand\bibinfo[2]{#2}
\providecommand\natexlab[1]{#1}
\providecommand\showeprint[2][]{arXiv:#2}

\bibitem[Chang et~al\mbox{.}(2024)]%
        {chang_plug-and-play_2024}
\bibfield{author}{\bibinfo{person}{Xiangyu Chang}, \bibinfo{person}{Sk~Miraj Ahmed}, \bibinfo{person}{Srikanth~V. Krishnamurthy}, \bibinfo{person}{Basak Guler}, \bibinfo{person}{Ananthram Swami}, \bibinfo{person}{Samet Oymak}, {and} \bibinfo{person}{Amit~K. Roy-Chowdhury}.} \bibinfo{year}{2024}\natexlab{}.
\newblock \bibinfo{title}{Plug-and-{Play} {Transformer} {Modules} for {Test}-{Time} {Adaptation}}.
\newblock
\newblock
\urldef\tempurl%
\url{http://arxiv.org/abs/2401.04130}
\showURL{%
\tempurl}
\newblock
\shownote{arXiv:2401.04130 [cs]}.


\bibitem[Chen et~al\mbox{.}(2022)]%
        {chen_contrastive_2022}
\bibfield{author}{\bibinfo{person}{Dian Chen}, \bibinfo{person}{Dequan Wang}, \bibinfo{person}{Trevor Darrell}, {and} \bibinfo{person}{Sayna Ebrahimi}.} \bibinfo{year}{2022}\natexlab{}.
\newblock \bibinfo{title}{Contrastive {Test}-{Time} {Adaptation}}.
\newblock
\newblock
\urldef\tempurl%
\url{http://arxiv.org/abs/2204.10377}
\showURL{%
\tempurl}
\newblock
\shownote{arXiv:2204.10377 [cs]}.


\bibitem[Choi et~al\mbox{.}(2021)]%
        {choi2021robustnet}
\bibfield{author}{\bibinfo{person}{Sungha Choi}, \bibinfo{person}{Sanghun Jung}, \bibinfo{person}{Huiwon Yun}, \bibinfo{person}{Joanne~T Kim}, \bibinfo{person}{Seungryong Kim}, {and} \bibinfo{person}{Jaegul Choo}.} \bibinfo{year}{2021}\natexlab{}.
\newblock \showarticletitle{Robustnet: Improving domain generalization in urban-scene segmentation via instance selective whitening}. In \bibinfo{booktitle}{\emph{Proceedings of the IEEE/CVF Conference on Computer Vision and Pattern Recognition}}. \bibinfo{pages}{11580--11590}.
\newblock


\bibitem[Gao et~al\mbox{.}(2023)]%
        {gao_back_2023}
\bibfield{author}{\bibinfo{person}{Jin Gao}, \bibinfo{person}{Jialing Zhang}, \bibinfo{person}{Xihui Liu}, \bibinfo{person}{Trevor Darrell}, \bibinfo{person}{Evan Shelhamer}, {and} \bibinfo{person}{Dequan Wang}.} \bibinfo{year}{2023}\natexlab{}.
\newblock \bibinfo{title}{Back to the {Source}: {Diffusion}-{Driven} {Test}-{Time} {Adaptation}}.
\newblock
\newblock
\urldef\tempurl%
\url{http://arxiv.org/abs/2207.03442}
\showURL{%
\tempurl}
\newblock
\shownote{arXiv:2207.03442 [cs]}.


\bibitem[Gong et~al\mbox{.}(2023)]%
        {gongNOTERobustContinual2023}
\bibfield{author}{\bibinfo{person}{Taesik Gong}, \bibinfo{person}{Jongheon Jeong}, \bibinfo{person}{Taewon Kim}, \bibinfo{person}{Yewon Kim}, \bibinfo{person}{Jinwoo Shin}, {and} \bibinfo{person}{Sung-Ju Lee}.} \bibinfo{year}{2023}\natexlab{}.
\newblock \bibinfo{title}{{{NOTE}}: {{Robust Continual Test-time Adaptation Against Temporal Correlation}}}.
\newblock
\newblock
\showeprint[arxiv]{2208.05117}~[cs]


\bibitem[He et~al\mbox{.}(2015)]%
        {heDeepResidualLearning2015}
\bibfield{author}{\bibinfo{person}{Kaiming He}, \bibinfo{person}{Xiangyu Zhang}, \bibinfo{person}{Shaoqing Ren}, {and} \bibinfo{person}{Jian Sun}.} \bibinfo{year}{2015}\natexlab{}.
\newblock \bibinfo{title}{Deep {{Residual Learning}} for {{Image Recognition}}}.
\newblock
\newblock
\showeprint[arxiv]{1512.03385}~[cs]


\bibitem[Hendrycks and Dietterich(2019)]%
        {hendrycks2019benchmarking}
\bibfield{author}{\bibinfo{person}{Dan Hendrycks} {and} \bibinfo{person}{Thomas Dietterich}.} \bibinfo{year}{2019}\natexlab{}.
\newblock \showarticletitle{Benchmarking neural network robustness to common corruptions and perturbations}.
\newblock \bibinfo{journal}{\emph{arXiv preprint arXiv:1903.12261}} (\bibinfo{year}{2019}).
\newblock


\bibitem[Ioffe and Szegedy(2015)]%
        {ioffeBatchNormalizationAccelerating2015}
\bibfield{author}{\bibinfo{person}{Sergey Ioffe} {and} \bibinfo{person}{Christian Szegedy}.} \bibinfo{year}{2015}\natexlab{}.
\newblock \bibinfo{title}{Batch {{Normalization}}: {{Accelerating Deep Network Training}} by {{Reducing Internal Covariate Shift}}}.
\newblock
\newblock
\showeprint[arxiv]{1502.03167}~[cs]


\bibitem[LEARNING({[n.\,d.]})]%
        {learningdataset}
\bibfield{author}{\bibinfo{person}{TASET SHIFT IN~MACHINE LEARNING}.} \bibinfo{year}{[n.\,d.]}\natexlab{}.
\newblock \showarticletitle{DATASET SHIFT IN MACHINE LEARNING}.
\newblock  (\bibinfo{year}{[n.\,d.]}).
\newblock


\bibitem[Lee et~al\mbox{.}(2024)]%
        {lee2024entropy}
\bibfield{author}{\bibinfo{person}{Jonghyun Lee}, \bibinfo{person}{Dahuin Jung}, \bibinfo{person}{Saehyung Lee}, \bibinfo{person}{Junsung Park}, \bibinfo{person}{Juhyeon Shin}, \bibinfo{person}{Uiwon Hwang}, {and} \bibinfo{person}{Sungroh Yoon}.} \bibinfo{year}{2024}\natexlab{}.
\newblock \showarticletitle{Entropy is not enough for test-time adaptation: From the perspective of disentangled factors}.
\newblock \bibinfo{journal}{\emph{arXiv preprint arXiv:2403.07366}} (\bibinfo{year}{2024}).
\newblock


\bibitem[Lim et~al\mbox{.}(2023)]%
        {limTTNDomainShiftAware2023}
\bibfield{author}{\bibinfo{person}{Hyesu Lim}, \bibinfo{person}{Byeonggeun Kim}, \bibinfo{person}{Jaegul Choo}, {and} \bibinfo{person}{Sungha Choi}.} \bibinfo{year}{2023}\natexlab{}.
\newblock \bibinfo{title}{{{TTN}}: {{A Domain-Shift Aware Batch Normalization}} in {{Test-Time Adaptation}}}.
\newblock
\newblock
\showeprint[arxiv]{2302.05155}~[cs]


\bibitem[Liu et~al\mbox{.}(2023)]%
        {liu2023vida}
\bibfield{author}{\bibinfo{person}{Jiaming Liu}, \bibinfo{person}{Senqiao Yang}, \bibinfo{person}{Peidong Jia}, \bibinfo{person}{Ming Lu}, \bibinfo{person}{Yandong Guo}, \bibinfo{person}{Wei Xue}, {and} \bibinfo{person}{Shanghang Zhang}.} \bibinfo{year}{2023}\natexlab{}.
\newblock \showarticletitle{Vida: Homeostatic visual domain adapter for continual test time adaptation}.
\newblock \bibinfo{journal}{\emph{arXiv preprint arXiv:2306.04344}} (\bibinfo{year}{2023}).
\newblock


\bibitem[McInnes et~al\mbox{.}(2017)]%
        {mcinnes2017hdbscan}
\bibfield{author}{\bibinfo{person}{Leland McInnes}, \bibinfo{person}{John Healy}, \bibinfo{person}{Steve Astels}, {et~al\mbox{.}}} \bibinfo{year}{2017}\natexlab{}.
\newblock \showarticletitle{hdbscan: Hierarchical density based clustering.}
\newblock \bibinfo{journal}{\emph{J. Open Source Softw.}} \bibinfo{volume}{2}, \bibinfo{number}{11} (\bibinfo{year}{2017}), \bibinfo{pages}{205}.
\newblock


\bibitem[Mirza et~al\mbox{.}(2022)]%
        {mirza2022norm}
\bibfield{author}{\bibinfo{person}{M~Jehanzeb Mirza}, \bibinfo{person}{Jakub Micorek}, \bibinfo{person}{Horst Possegger}, {and} \bibinfo{person}{Horst Bischof}.} \bibinfo{year}{2022}\natexlab{}.
\newblock \showarticletitle{The norm must go on: Dynamic unsupervised domain adaptation by normalization}. In \bibinfo{booktitle}{\emph{Proceedings of the IEEE/CVF conference on computer vision and pattern recognition}}. \bibinfo{pages}{14765--14775}.
\newblock


\bibitem[Murtagh and Legendre(2014)]%
        {murtagh2014ward}
\bibfield{author}{\bibinfo{person}{Fionn Murtagh} {and} \bibinfo{person}{Pierre Legendre}.} \bibinfo{year}{2014}\natexlab{}.
\newblock \showarticletitle{Ward’s hierarchical agglomerative clustering method: which algorithms implement Ward’s criterion?}
\newblock \bibinfo{journal}{\emph{Journal of classification}}  \bibinfo{volume}{31} (\bibinfo{year}{2014}), \bibinfo{pages}{274--295}.
\newblock


\bibitem[Neyshabur et~al\mbox{.}(2020)]%
        {neyshabur2020being}
\bibfield{author}{\bibinfo{person}{Behnam Neyshabur}, \bibinfo{person}{Hanie Sedghi}, {and} \bibinfo{person}{Chiyuan Zhang}.} \bibinfo{year}{2020}\natexlab{}.
\newblock \showarticletitle{What is being transferred in transfer learning?}
\newblock \bibinfo{journal}{\emph{Advances in neural information processing systems}}  \bibinfo{volume}{33} (\bibinfo{year}{2020}), \bibinfo{pages}{512--523}.
\newblock


\bibitem[Niu et~al\mbox{.}({[n.\,d.]})]%
        {niuEfficientTestTimeModel}
\bibfield{author}{\bibinfo{person}{Shuaicheng Niu}, \bibinfo{person}{Jiaxiang Wu}, \bibinfo{person}{Yifan Zhang}, \bibinfo{person}{Yaofo Chen}, \bibinfo{person}{Shijian Zheng}, \bibinfo{person}{Peilin Zhao}, {and} \bibinfo{person}{Mingkui Tan}.} \bibinfo{year}{[n.\,d.]}\natexlab{}.
\newblock \showarticletitle{Efficient {{Test-Time Model Adaptation}} without {{Forgetting}}}.
\newblock  (\bibinfo{year}{[n.\,d.]}).
\newblock


\bibitem[Niu et~al\mbox{.}(2022)]%
        {niu2022efficient}
\bibfield{author}{\bibinfo{person}{Shuaicheng Niu}, \bibinfo{person}{Jiaxiang Wu}, \bibinfo{person}{Yifan Zhang}, \bibinfo{person}{Yaofo Chen}, \bibinfo{person}{Shijian Zheng}, \bibinfo{person}{Peilin Zhao}, {and} \bibinfo{person}{Mingkui Tan}.} \bibinfo{year}{2022}\natexlab{}.
\newblock \showarticletitle{Efficient Test-Time Model Adaptation without Forgetting}. In \bibinfo{booktitle}{\emph{The Internetional Conference on Machine Learning}}.
\newblock


\bibitem[Niu et~al\mbox{.}(2023a)]%
        {niuSTABLETESTTIMEADAPTATION2023}
\bibfield{author}{\bibinfo{person}{Shuaicheng Niu}, \bibinfo{person}{Jiaxiang Wu}, \bibinfo{person}{Yifan Zhang}, \bibinfo{person}{Zhiquan Wen}, \bibinfo{person}{Yaofo Chen}, \bibinfo{person}{Peilin Zhao}, {and} \bibinfo{person}{Mingkui Tan}.} \bibinfo{year}{2023}\natexlab{a}.
\newblock \showarticletitle{{{TOWARDS STABLE TEST-TIME ADAPTATION IN DYNAMIC WILD WORLD}}}.
\newblock  (\bibinfo{year}{2023}).
\newblock


\bibitem[Niu et~al\mbox{.}(2023b)]%
        {niu2023towards}
\bibfield{author}{\bibinfo{person}{Shuaicheng Niu}, \bibinfo{person}{Jiaxiang Wu}, \bibinfo{person}{Yifan Zhang}, \bibinfo{person}{Zhiquan Wen}, \bibinfo{person}{Yaofo Chen}, \bibinfo{person}{Peilin Zhao}, {and} \bibinfo{person}{Mingkui Tan}.} \bibinfo{year}{2023}\natexlab{b}.
\newblock \showarticletitle{Towards stable test-time adaptation in dynamic wild world}.
\newblock \bibinfo{journal}{\emph{arXiv preprint arXiv:2302.12400}} (\bibinfo{year}{2023}).
\newblock


\bibitem[Pan et~al\mbox{.}(2018)]%
        {pan2018two}
\bibfield{author}{\bibinfo{person}{Xingang Pan}, \bibinfo{person}{Ping Luo}, \bibinfo{person}{Jianping Shi}, {and} \bibinfo{person}{Xiaoou Tang}.} \bibinfo{year}{2018}\natexlab{}.
\newblock \showarticletitle{Two at once: Enhancing learning and generalization capacities via ibn-net}. In \bibinfo{booktitle}{\emph{Proceedings of the european conference on computer vision (ECCV)}}. \bibinfo{pages}{464--479}.
\newblock


\bibitem[Recht et~al\mbox{.}(2019)]%
        {recht2019imagenet}
\bibfield{author}{\bibinfo{person}{Benjamin Recht}, \bibinfo{person}{Rebecca Roelofs}, \bibinfo{person}{Ludwig Schmidt}, {and} \bibinfo{person}{Vaishaal Shankar}.} \bibinfo{year}{2019}\natexlab{}.
\newblock \showarticletitle{Do imagenet classifiers generalize to imagenet?}. In \bibinfo{booktitle}{\emph{International conference on machine learning}}. PMLR, \bibinfo{pages}{5389--5400}.
\newblock


\bibitem[Sander et~al\mbox{.}(1998)]%
        {sander1998density}
\bibfield{author}{\bibinfo{person}{J{\"o}rg Sander}, \bibinfo{person}{Martin Ester}, \bibinfo{person}{Hans-Peter Kriegel}, {and} \bibinfo{person}{Xiaowei Xu}.} \bibinfo{year}{1998}\natexlab{}.
\newblock \showarticletitle{Density-based clustering in spatial databases: The algorithm gdbscan and its applications}.
\newblock \bibinfo{journal}{\emph{Data mining and knowledge discovery}}  \bibinfo{volume}{2} (\bibinfo{year}{1998}), \bibinfo{pages}{169--194}.
\newblock


\bibitem[Sarfraz et~al\mbox{.}(2019)]%
        {sarfraz2019efficient}
\bibfield{author}{\bibinfo{person}{Saquib Sarfraz}, \bibinfo{person}{Vivek Sharma}, {and} \bibinfo{person}{Rainer Stiefelhagen}.} \bibinfo{year}{2019}\natexlab{}.
\newblock \showarticletitle{Efficient parameter-free clustering using first neighbor relations}. In \bibinfo{booktitle}{\emph{Proceedings of the IEEE/CVF conference on computer vision and pattern recognition}}. \bibinfo{pages}{8934--8943}.
\newblock


\bibitem[Wang et~al\mbox{.}(2021)]%
        {wangTentFullyTesttime2021b}
\bibfield{author}{\bibinfo{person}{Dequan Wang}, \bibinfo{person}{Evan Shelhamer}, \bibinfo{person}{Shaoteng Liu}, \bibinfo{person}{Bruno Olshausen}, {and} \bibinfo{person}{Trevor Darrell}.} \bibinfo{year}{2021}\natexlab{}.
\newblock \bibinfo{title}{Tent: {{Fully Test-time Adaptation}} by {{Entropy Minimization}}}.
\newblock
\newblock
\showeprint[arxiv]{2006.10726}~[cs, stat]


\bibitem[Wang et~al\mbox{.}(2024)]%
        {wang_decoupled_2024}
\bibfield{author}{\bibinfo{person}{Guowei Wang}, \bibinfo{person}{Changxing Ding}, \bibinfo{person}{Wentao Tan}, {and} \bibinfo{person}{Mingkui Tan}.} \bibinfo{year}{2024}\natexlab{}.
\newblock \bibinfo{title}{Decoupled {Prototype} {Learning} for {Reliable} {Test}-{Time} {Adaptation}}.
\newblock
\newblock
\urldef\tempurl%
\url{http://arxiv.org/abs/2401.08703}
\showURL{%
\tempurl}
\newblock
\shownote{arXiv:2401.08703 [cs]}.


\bibitem[Wang et~al\mbox{.}(2022)]%
        {wangContinualTestTimeDomain2022}
\bibfield{author}{\bibinfo{person}{Qin Wang}, \bibinfo{person}{Olga Fink}, \bibinfo{person}{Luc Van~Gool}, {and} \bibinfo{person}{Dengxin Dai}.} \bibinfo{year}{2022}\natexlab{}.
\newblock \bibinfo{title}{Continual {{Test-Time Domain Adaptation}}}.
\newblock
\newblock
\showeprint[arxiv]{2203.13591}~[cs]


\bibitem[Wang et~al\mbox{.}(2023)]%
        {wang2023dynamically}
\bibfield{author}{\bibinfo{person}{Wei Wang}, \bibinfo{person}{Zhun Zhong}, \bibinfo{person}{Weijie Wang}, \bibinfo{person}{Xi Chen}, \bibinfo{person}{Charles Ling}, \bibinfo{person}{Boyu Wang}, {and} \bibinfo{person}{Nicu Sebe}.} \bibinfo{year}{2023}\natexlab{}.
\newblock \showarticletitle{Dynamically Instance-Guided Adaptation: A Backward-Free Approach for Test-Time Domain Adaptive Semantic Segmentation}. In \bibinfo{booktitle}{\emph{Proceedings of the IEEE/CVF Conference on Computer Vision and Pattern Recognition}}. \bibinfo{pages}{24090--24099}.
\newblock


\bibitem[Yuan et~al\mbox{.}(2023)]%
        {yuanRobustTestTimeAdaptation2023}
\bibfield{author}{\bibinfo{person}{Longhui Yuan}, \bibinfo{person}{Binhui Xie}, {and} \bibinfo{person}{Shuang Li}.} \bibinfo{year}{2023}\natexlab{}.
\newblock \bibinfo{title}{Robust {{Test-Time Adaptation}} in {{Dynamic Scenarios}}}.
\newblock
\newblock
\showeprint[arxiv]{2303.13899}~[cs]


\bibitem[Zhang et~al\mbox{.}(2022)]%
        {zhang2022memo}
\bibfield{author}{\bibinfo{person}{Marvin Zhang}, \bibinfo{person}{Sergey Levine}, {and} \bibinfo{person}{Chelsea Finn}.} \bibinfo{year}{2022}\natexlab{}.
\newblock \showarticletitle{Memo: Test time robustness via adaptation and augmentation}.
\newblock \bibinfo{journal}{\emph{Advances in neural information processing systems}}  \bibinfo{volume}{35} (\bibinfo{year}{2022}), \bibinfo{pages}{38629--38642}.
\newblock


\bibitem[Zhang et~al\mbox{.}(1996)]%
        {zhang1996birch}
\bibfield{author}{\bibinfo{person}{Tian Zhang}, \bibinfo{person}{Raghu Ramakrishnan}, {and} \bibinfo{person}{Miron Livny}.} \bibinfo{year}{1996}\natexlab{}.
\newblock \showarticletitle{BIRCH: an efficient data clustering method for very large databases}.
\newblock \bibinfo{journal}{\emph{ACM sigmod record}} \bibinfo{volume}{25}, \bibinfo{number}{2} (\bibinfo{year}{1996}), \bibinfo{pages}{103--114}.
\newblock


\bibitem[Zhao et~al\mbox{.}(2023)]%
        {zhao2023ttab}
\bibfield{author}{\bibinfo{person}{Hao Zhao}, \bibinfo{person}{Yuejiang Liu}, \bibinfo{person}{Alexandre Alahi}, {and} \bibinfo{person}{Tao Lin}.} \bibinfo{year}{2023}\natexlab{}.
\newblock \showarticletitle{On Pitfalls of Test-time Adaptation}. In \bibinfo{booktitle}{\emph{International Conference on Machine Learning (ICML)}}.
\newblock


\bibitem[Zhou et~al\mbox{.}(2024)]%
        {zhou_resilient_2024}
\bibfield{author}{\bibinfo{person}{Xingzhi Zhou}, \bibinfo{person}{Zhiliang Tian}, \bibinfo{person}{Ka~Chun Cheung}, \bibinfo{person}{Simon See}, {and} \bibinfo{person}{Nevin~L. Zhang}.} \bibinfo{year}{2024}\natexlab{}.
\newblock \bibinfo{title}{Resilient {Practical} {Test}-{Time} {Adaptation}: {Soft} {Batch} {Normalization} {Alignment} and {Entropy}-driven {Memory} {Bank}}.
\newblock
\newblock
\urldef\tempurl%
\url{http://arxiv.org/abs/2401.14619}
\showURL{%
\tempurl}
\newblock
\shownote{arXiv:2401.14619 [cs]}.


\end{thebibliography}

\end{document}


\title{Supplementary Materials: Discover Your Neighbors: Advanced Stable Test-Time Adaptation in Dynamic World}

\maketitle
\appendix

\section{The Layer-wise Instance Statistics Clustering Algorithm}
Algorithm~\ref{algorithm_LISC} is the proposed Layer-wise Instance Statistics Clustering (LISC) Algorithm in this paper.
By calculating the cosine similarity between the feature maps of two samples at a certain BN layer, the first neighbor with the most similar feature distribution can be identified precisely. Then by connecting each sample to its first neighbor and the first neighbor's neighbors through the adjacency matrix in equation 4, multiple clusters can be formed rapidly. This enables the feature maps of different samples within each cluster to provide proximate class-irrelevant feature distributions. In addition, we observed that there are significantly more clusters in the shallow layers of the model compared to the deeper layers. This implies that the earlier layers of the model focus more on learning class-irrelevant features that vary across domains, while the deeper layers learn more class-relevant features that are invariant to domain shifts.

\begin{algorithm}[t]
\caption{LISC Algorithm}\label{algo:the_alg}
\begin{algorithmic}[1]
\STATE \textbf{Input:}  Feature map $F_{j}=\{1,2,\cdots,B\}$ in the BN layer $j$, $F_{j} \in \mathbb{R}^{ B \times C \times H \times W}$, $1 \le j \le N$, where $N$ is total number of BN layers in the model.
\STATE \textbf{Output:} Feature map $F_{j}^{'}=\{C_{1},C_{2},\cdots,C_{k}\}$ after clustering, $C_{i} \in \mathbb{R}^{ b \times C \times H \times W}$, $1\le i\le k$.
\STATE \textbf{Begin LISC Algorithm:}
\STATE Compute first neighbors integer vector $n^1 \in \mathbb{R}^{ B \times 1} $ via exact distance (the distance metric is defined by equation 3).
\STATE  Given $n^1$ get clusters of the feature map $F_{j}$ via equation 4. 

\STATE \textbf{END}
\end{algorithmic}
\label{algorithm_LISC}
\end{algorithm}

\section{Details of Datasets}
\begin{figure*}[htbp]   
\centering
\includegraphics[width=1\textwidth]{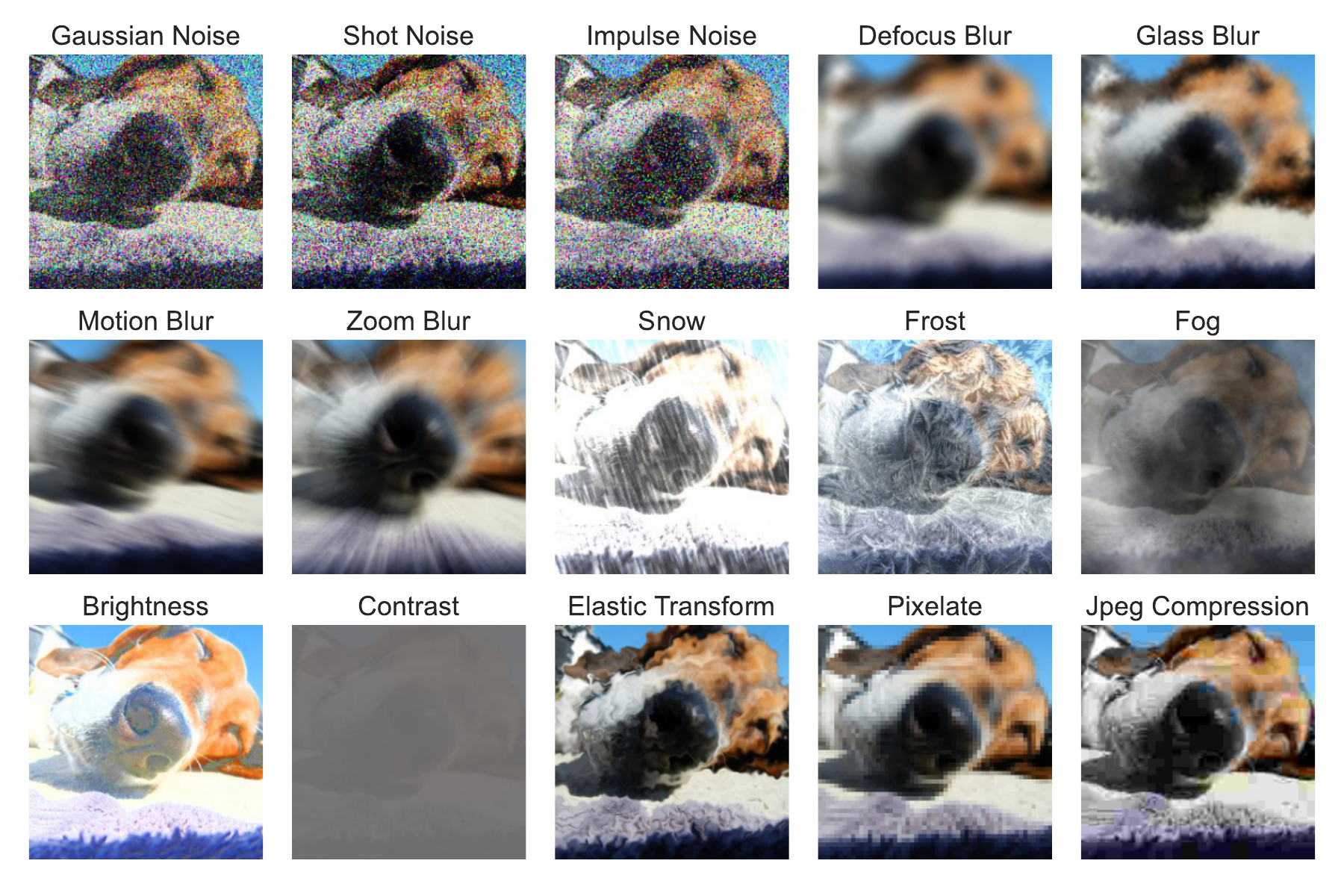}
                   \caption{
                   Visualization of the 15 types of corruptions in CIFAR10-C, CIFAR100-C and ImageNet-C.}
    \label{fig:dog_in_15corruptions}
\end{figure*}

CIFAR10-C, CIFAR100-C, and ImageNet-C are three commonly used TTA benchmarks to measure model robustness under covariate shift. Each type of corruption has 5 severity levels, where a higher level indicates a greater degree of data drift. CIFAR10-C and CIFAR100-C  comprise 10 and 100 classes respectively, with 50,000 training samples and 10,000 test samples.  ImageNet-C is a large-scale dataset with 1,000 classes, consisting of 1,281,167 training samples and 50,000 test samples. As illustrated in figure~\ref{fig:dog_in_15corruptions}, each of the three datasets contains 15 types of corruptions, which are: Gaussian Noise,
Shot Noise, Impulse Noise, Defocus Blur, Glass Blur, Motion Blur, Zoom Blur, Snow, Frost, Fog, Brightness, Contrast, Elastic Transformation, Pixelate, and JPEG Compression.

\section{Hyperparameter  Settings}
The hyperparameters can be divided into two types: one type is shared by all the baselines, and the other type consists of hyperparameters specific to each method. For the shared hyperparameters, $batch \  size = 64$, $learning\ rate = 1e^{-4} $. The optimizer used is SGD. The hyperparameters of each test-time fine-tuning method are set according to the TTAB benchmark [32] (the optimal hyperparameters that achieved the best performance for each method in the original paper). Following are the hyperparameters specific to each test-time normalization method:
\begin{itemize}
\item The hyperparameters of TBN are set accrording to the settings in [25];
\item The hyperparameters of IABN are set accrording to the settings in [5]; 
\item The hyperparameters of $\alpha$-BN are set accrording to the settings in [25].
\end{itemize}

\section{Experimental Results and Analysis under
Wild Scenario}

The Wild scenario is obtained by adding a label shift to the Random scenario. Specifically, in the Wild scenario, samples in a batch may come from the same distribution/different distributions, while label shift may also exist. This scenario maximally simulates the real data flow in the open wild world, hence referred to as 'Wild'. Similar to \textbf{NOTE} [5], we utilize the Dirichlet distribution to construct temporally correlated streams (label shift). The concentration parameter $\delta$ ($\delta > 0$) of the Dirichlet distribution controls the degree of label shift. Smaller $\delta$ leads to larger shifts. Here we set $\delta=0.1, \delta=0.01$, and $\delta=0.005$ respectively. Table~\ref{non_iidness = 0.1}, table~\ref{non_iidness = 0.01}, and table~\ref{non_iidness = 0.005} present the corresponding results. These tables demonstrate that as the degree of label shift increases, the performance of our method barely declines, exhibiting robustness under label shift. Additionally, our method achieves approximately 5\% higher average accuracy compared to the best performing baseline, and around 20\% higher accuracy relative to the poorest performing baseline. In contrast, DeYo exhibits deteriorating performance with increasing label shifts (accuracy below 20\% on CIFAR100-C), and other methods also suffer noticeable performance drops. This shows that when facing wild data streams in an open world, our method maintains high performance and robustness, with negligible impact from data stream distribution shifts. This far surpasses existing methods in handling such challenging real-world data streams.

\section{Experimental Results and Analysis on Simulated Life-Long Adaptation under CrossMix Scenario}

Currently, baseline methods are designed on the basis of continual TTA, meaning models need to adapt continuously to maintain performance when facing non-station incoming data streams over long periods. In previous works, experiments simulate this continual learning setting by feeding the full dataset sequentially into models for inference and adaptation, tracking average accuracy over the process. However, in real-world deployment, inference cycles are much longer. Therefore, to better simulate continual TTA in real environments, we replicate datasets multiple times to obtain longer datasets and thus longer inference cycles. Here, we duplicate CIFAR10-C and CIFAR100-C two and three times respectively ($Round = 2$ and $Round = 3$). Table~\ref{rounds = 2} and table~\ref{rounds = 3} present the corresponding results (CrossMix scenario).
The two tables show that our method exhibits negligible accuracy drop from Round=2 to Round=3. In contrast, from Round=2 to Round=3, DeYo's accuracy decreases by 2\% on CIFAR10-C and 5\% on CIFAR100-C. TENT sees a 3\% accuracy decline on CIFAR10-C and 4\% on CIFAR100-C. SAR's accuracy drops by 3\% on CIFAR10-C and 3\% on CIFAR100-C. Other methods also suffer varying degrees of accuracy degradation. This demonstrates that as the cycle of continual TTA lengthens, other methods exhibit deteriorating performance, while our method maintains consistent stability.

\begin{table*}[htbp]
    \caption{Comparisons with state-of-the-art methods on CIFAR10-C and CIFAR100-C respectively (severity level = 5) under \textbf{\textsc{Batch Size=64}} regarding \textbf{Accuracy (\%)}. Each method was evaluated under the Wild scenario ($\delta = 0.1$) with using a ResNet-50 model architecture. The best result is denoted in bold black font.
    }
    \label{tab:imagenet-c-bs1-level3}
\newcommand{\tabincell}[2]{\begin{tabular}{@{}#1@{}}#2\end{tabular}}
 \begin{center}
 \begin{threeparttable}
  \scriptsize
    \resizebox{0.5\linewidth}{!}{
  \begin{tabular}{l|cc>{\columncolor{blue!8}}c}
  

  Method & CIFAR10-C & CIFAR100-C & Avg.\\

  \midrule
  Source (ResNet-50)& 57.40 & 28.59 &  43.00 \\
   \rowcolor[HTML]{d0d0d0}
         \multicolumn{4}{c}{\scriptsize\textbf{\textsc{Test-Time Fine-Tune}}} \\
         ~~$\bullet~$EATA & 54.99 & 35.60 &  45.30 \\ 
        ~~$\bullet~$DeYO & 50.77 & 18.27 & 34.52 \\
        ~~$\bullet~$SAR    & 52.19 & 32.99 & 42.59 \\

        ~~$\bullet~$TENT   & 52.48 & 34.58 & 43.53 \\
        ~~$\bullet~$NOTE & 67.50 & 24.70 &  46.10\\
        ~~$\bullet~$ViDA & 55.02 & 36.72 & 45.87 \\ 
       
         ~~$\bullet~$RoTTA & 49.49 & 22.60 & 36.05 \\

 \rowcolor[HTML]{d0d0d0}
         \multicolumn{4}{c}{\scriptsize\textbf{\textsc{Test-Time Normalization}}} \\

       ~~$\bullet~$TBN  & 55.00 & 36.73 &  45.87\\
       ~~$\bullet~$$\alpha$-BN   & 60.98 & \textbf{40.51} & 50.75 \\
       ~~$\bullet~$IABN    & 63.30 & 24.92 & 44.11 \\
      
       \rowcolor{pink!30}~~$\bullet~$DYN (ours)  & \textbf{69.49} & 38.86 & \textbf{54.18} \\
        
 \end{tabular}
 }
  \end{threeparttable}
  \end{center}
  \label{non_iidness = 0.1}
\end{table*}

\begin{table*}[htbp]
    \caption{Comparisons with state-of-the-art methods on CIFAR10-C and CIFAR100-C respectively (severity level = 5) under \textbf{\textsc{Batch Size=64}} regarding \textbf{Accuracy (\%)}. Each method was evaluated under the Wild scenario ($\delta = 0.01$) with using a ResNet-50 model architecture. The best result  is denoted in bold black font.
    }
    \label{tab:imagenet-c-bs1-level3}
\newcommand{\tabincell}[2]{\begin{tabular}{@{}#1@{}}#2\end{tabular}}
 \begin{center}
 \begin{threeparttable}
  \scriptsize
    \resizebox{0.5\linewidth}{!}{
  \begin{tabular}{l|cc>{\columncolor{blue!8}}c}
  

  Method & CIFAR10-C & CIFAR100-C & Avg.\\

  \midrule
  Source (ResNet-50)& 57.40 & 28.59 &  43.00 \\
   \rowcolor[HTML]{d0d0d0}
         \multicolumn{4}{c}{\scriptsize\textbf{\textsc{Test-Time Fine-Tune}}} \\
         ~~$\bullet~$EATA & 54.47 & 24.39 &  39.43\\
        ~~$\bullet~$DeYO & 50.35 & 11.27 & 30.81 \\
        ~~$\bullet~$SAR    & 51.48 & 23.34 & 37.41 \\

        ~~$\bullet~$TENT   & 51.77 & 27.16 & 39.47 \\
        ~~$\bullet~$NOTE & 67.49 &  24.71     &  46.10\\
        ~~$\bullet~$ViDA & 54.47 & 30.59 &   42.53\\ 
       
         ~~$\bullet~$RoTTA & 49.27 & 22.33 &  35.80 \\

 \rowcolor[HTML]{d0d0d0}
         \multicolumn{4}{c}{\scriptsize\textbf{\textsc{Test-Time Normalization}}} \\

       ~~$\bullet~$TBN & 54.47 & 30.57 & 42.52 \\
       ~~$\bullet~$$\alpha$-BN   & 60.54 & 34.51 & 47.53 \\
       ~~$\bullet~$IABN    & 63.28 & 24.97 &  44.13 \\
      
       \rowcolor{pink!30}~~$\bullet~$DYN (ours)  & \textbf{69.43} & \textbf{35.56} &  \textbf{52.50}\\
        
 \end{tabular}
 }
  \end{threeparttable}
  \end{center}
  \label{non_iidness = 0.01}
\end{table*}

\begin{table*}[htbp]
    \caption{Comparisons with state-of-the-art methods on CIFAR10-C and CIFAR100-C respectively (severity level = 5) under \textbf{\textsc{Batch Size=64}} regarding \textbf{Accuracy (\%)}. Each method was evaluated under the Wild scenario ($\delta = 0.005$) with using a ResNet-50 model architecture. The best result  is denoted in bold black font.
    }
    \label{tab:imagenet-c-bs1-level3}
\newcommand{\tabincell}[2]{\begin{tabular}{@{}#1@{}}#2\end{tabular}}
 \begin{center}
 \begin{threeparttable}
  \scriptsize
    \resizebox{0.5\linewidth}{!}{
  \begin{tabular}{l|cc>{\columncolor{blue!8}}c}
  

  Method & CIFAR10-C & CIFAR100-C & Avg.\\

  \midrule
  Source (ResNet-50) & 57.40 & 28.59 &  43.00 \\
   \rowcolor[HTML]{d0d0d0}
         \multicolumn{4}{c}{\scriptsize\textbf{\textsc{Test-Time Fine-Tune}}} \\
         ~~$\bullet~$EATA  & 54.42 & 23.42 & 38.92  \\
        ~~$\bullet~$DeYO & 48.04 & 10.17 & 29.11   \\
        ~~$\bullet~$SAR    & 51.45 & 22.34 &  36.90  \\

        ~~$\bullet~$TENT    & 51.99 & 25.97 & 38.98   \\
        ~~$\bullet~$NOTE & 67.56 & 24.70 &  46.13\\
        ~~$\bullet~$ViDA & 54.42 & 29.81 & 42.12 \\ 
       
         ~~$\bullet~$RoTTA  & 49.14 & 21.87 & 35.51 \\

 \rowcolor[HTML]{d0d0d0}
         \multicolumn{4}{c}{\scriptsize\textbf{\textsc{Test-Time Normalization}}} \\

       ~~$\bullet~$TBN & 54.42 & 29.80 & 42.11 \\
       ~~$\bullet~$$\alpha$-BN    & 60.46 & 33.52 & 46.99 \\
       ~~$\bullet~$IABN    & 63.30 & 24.95 & 44.12 \\
      
       \rowcolor{pink!30}~~$\bullet~$DYN (ours)  & \textbf{69.38} & \textbf{35.21} & \textbf{52.30} \\
        
 \end{tabular}
 }
  \end{threeparttable}
  \end{center}
  \label{non_iidness = 0.005}
\end{table*}

\begin{table*}[htbp]
    \caption{Comparisons with state-of-the-art methods on CIFAR10-C and CIFAR100-C respectively (severity level = 5) under \textbf{\textsc{Batch Size=64}} regarding \textbf{Accuracy (\%)}. Each method was evaluated under the CorssMix scenario ($Round = 2$) with using a ResNet-50 model architecture. The best result  is denoted in bold black font.
    }
    \label{tab:imagenet-c-bs1-level3}
\newcommand{\tabincell}[2]{\begin{tabular}{@{}#1@{}}#2\end{tabular}}
 \begin{center}
 \begin{threeparttable}
  \scriptsize
    \resizebox{0.5\linewidth}{!}{
  \begin{tabular}{l|cc>{\columncolor{blue!8}}c}
  

  Method & CIFAR10-C & CIFAR100-C & Avg.\\

  \midrule
  Source (ResNet-50)& 57.41 & 28.59 & 43.00  \\
   \rowcolor[HTML]{d0d0d0}
         \multicolumn{4}{c}{\scriptsize\textbf{\textsc{Test-Time Fine-Tune}}} \\
         ~~$\bullet~$EATA & 62.03 & 34.38 &  48.20 \\
        ~~$\bullet~$DeYO  & 67.89 & 17.38 & 42.63 \\
        ~~$\bullet~$SAR    & 57.54 & 27.59 & 42.56 \\

        ~~$\bullet~$TENT   & 58.89 & 27.20 & 43.05 \\
        ~~$\bullet~$NOTE & 66.68 & 24.70 &  45.69 \\
        ~~$\bullet~$ViDA &  62.01    &  32.29     &47.15  \\
       
         ~~$\bullet~$RoTTA  &  49.11     &   23.58    &  36.34 \\

 \rowcolor[HTML]{d0d0d0}
         \multicolumn{4}{c}{\scriptsize\textbf{\textsc{Test-Time Normalization}}} \\

       ~~$\bullet~$TBN  & 62.01 & 32.29 & 47.15 \\
       ~~$\bullet~$$\alpha$-BN   & 62.45 & 33.21 &  47.83\\
       ~~$\bullet~$IABN    & 63.34 & 24.91 & 44.12 \\
      
       \rowcolor{pink!30}~~$\bullet~$DYN (ours)  & \textbf{71.48} &  \textbf{39.72} &  \textbf{55.60} \\
        
 \end{tabular}
 }
  \end{threeparttable}
  \end{center}
  \label{rounds = 2}
\end{table*}

\begin{table*}[htbp]
    \caption{Comparisons with state-of-the-art methods on CIFAR10-C and CIFAR100-C respectively (severity level = 5) under \textbf{\textsc{Batch Size=64}} regarding \textbf{Accuracy (\%)}. Each method was evaluated under the CorssMix scenario ($Round = 3$) with using a ResNet-50 model architecture. The best result  is denoted in bold black font.
    }
    \label{tab:imagenet-c-bs1-level3}
\newcommand{\tabincell}[2]{\begin{tabular}{@{}#1@{}}#2\end{tabular}}
 \begin{center}
 \begin{threeparttable}
  \scriptsize
    \resizebox{0.5\linewidth}{!}{
  \begin{tabular}{l|cc>{\columncolor{blue!8}}c}
  

  Method & CIFAR10-C & CIFAR100-C & Avg.\\

  \midrule
  Source (ResNet-50)& 57.39 & 28.59 & 42.99  \\
   \rowcolor[HTML]{d0d0d0}
         \multicolumn{4}{c}{\scriptsize\textbf{\textsc{Test-Time Fine-Tune}}} \\
         ~~$\bullet~$EATA & 62.01 & 33.67 &47.84  \\ 
        ~~$\bullet~$DeYO & 65.77 & 12.77 & 39.27 \\
        ~~$\bullet~$SAR    & 54.28 & 24.61 & 39.45 \\

        ~~$\bullet~$TENT   & 55.49 & 23.64 &  39.56\\
        ~~$\bullet~$NOTE & 67.90 & 24.70 &  46.30 \\ 
        ~~$\bullet~$ViDA & 62.02 & 32.21 & 47.12 \\
       
         ~~$\bullet~$RoTTA & 49.21 & 22.11 &  35.66 \\

 \rowcolor[HTML]{d0d0d0}
         \multicolumn{4}{c}{\scriptsize\textbf{\textsc{Test-Time Normalization}}} \\

       ~~$\bullet~$TBN & 62.02 & 32.21 &47.12  \\
       ~~$\bullet~$$\alpha$-BN   & 62.46 & 33.21 & 47.84 \\
       ~~$\bullet~$IABN    & 63.28 & 24.87 & 44.08 \\
      
       \rowcolor{pink!30}~~$\bullet~$DYN (ours)   & \textbf{71.54} & \textbf{39.67} & \textbf{55.61} \\
        
 \end{tabular}
 }
  \end{threeparttable}
  \end{center}
  \label{rounds = 3}
\end{table*}

